\PassOptionsToPackage{breaklinks}{hyperref}
\documentclass{article}

\usepackage[main, final]{neurips_2026}

\usepackage[utf8]{inputenc}
\usepackage[T1]{fontenc}
\usepackage{hyperref}
\usepackage{url}
\usepackage{booktabs}
\usepackage{amsfonts}
\usepackage{amsmath}
\usepackage{amssymb}
\usepackage{nicefrac}
\usepackage{microtype}
\usepackage[table]{xcolor}
\usepackage{graphicx}
\usepackage{algorithm}
\usepackage{algpseudocode}
\usepackage{multirow}
\usepackage{caption}
\usepackage{marvosym}
\usepackage{listings}
\usepackage[normalem]{ulem}
\usepackage{tabularx}
\usepackage{xspace}
\usepackage{cleveref}
\usepackage{siunitx}
\hypersetup{
    colorlinks=false,
    citebordercolor=green,
    linkbordercolor={1 1 1},
    urlbordercolor={1 1 1}
}

\newcommand{\methodname}{UltraFlash}
\newcommand{\basename}{PixelRush}

\newcommand{\eg}{\emph{e.g.}\xspace}

\definecolor{codegreen}{rgb}{0,0.6,0}
\definecolor{codegray}{rgb}{0.5,0.5,0.5}
\definecolor{codepurple}{rgb}{0.58,0,0.82}
\definecolor{backcolour}{rgb}{0.95,0.95,0.92}

\definecolor{mydarkblue}{rgb}{0,0.08,1}
\definecolor{mydarkgreen}{rgb}{0.02,0.6,0.02}
\definecolor{myred}{rgb}{1.0,0.0,0.0}
\definecolor{myred2}{rgb}{0.7,0.1,0.1}
\definecolor{mydarkblue2}{rgb}{0.05,0.1,0.7}
\definecolor{mypurple}{rgb}{111,0,255}
\definecolor{mypurple2}{rgb}{111,0,111}

\definecolor{dkgreen}{rgb}{0,0.6,0}
\definecolor{gray}{rgb}{0.5,0.5,0.5}
\definecolor{mauve}{rgb}{0.58,0,0.82}

\newcommand{\Loss}{\mathcal{L}}

\newcommand{\minisection}[1]{\noindent{\textbf{#1}}}
\newcommand{\myparagraph}[1]{\vspace{1mm}\noindent{\textit{#1}}}

\title{\methodname: Accelerating Megapixel Visual Synthesis}

\makeatletter
\let\ultraflash@fnsymbol\@fnsymbol
\renewcommand{\@fnsymbol}[1]{\ifnum#1=1 \ensuremath{\star}\else\ultraflash@fnsymbol{#1}\fi}
\makeatother

\author{%
  \href{https://laihongphuc.github.io/}{Phuc~Lai}\thanks{Equal contribution.}
  \qquad
  \href{https://aengusng8.github.io/}{Anh~Nguyen}\footnotemark[1]
  \qquad
  \href{https://phongnhhn.info/}{Phong~Nguyen}
  \qquad
  {\href{https://scholar.google.com/citations?user=FYZ5ODQAAAAJ&hl=en}{Anh~Tran}}
  \\
  Qualcomm AI Research\thanks{Qualcomm AI Research is an initiative of Qualcomm Technologies, Inc.}
}

\begin{document}

\maketitle

\begin{center}
    \includegraphics[width=\linewidth]{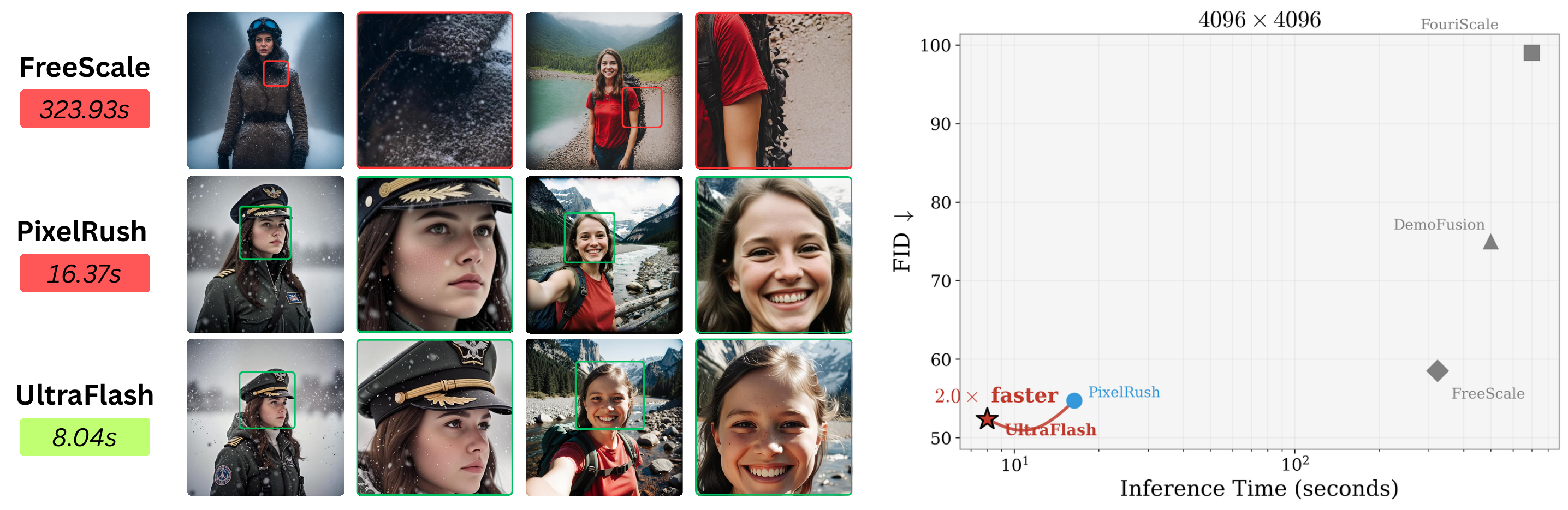}
    \captionof{figure}{\textbf{\methodname{} improves the 4K speed--quality frontier.} It achieves a \(2\times\) speedup over the current state-of-the-art cascade baseline with better FID, moving closer to the Pareto-optimal frontier.}
    \label{fig:teaser}
\end{center}

\begin{abstract}
Megapixel visual synthesis with latent diffusion models requires operating far beyond the resolution at which these models are trained. Cascade refinement is the most successful approach to handling this problem: it upgrades a base image at a regular resolution to the target resolution via multiple cascade upsampling steps, each involving a resolution upgrade traversed through pixel space and a denoising-based refinement in latent space. Although denoising is often treated as the main cost, we identify an underexplored source of overhead: repeatedly moving representations through the VAE. We refer to these steps as \emph{pixel-space transitions}: intermediate latents are decoded to RGB, resized, and re-encoded before refinement, while final latents are decoded patch by patch with dense overlap to suppress artifacts. We present \textbf{UltraFlash}, which shifts the focus of acceleration from denoising alone to the VAE transitions around it. Its first component, \textbf{Latent Hyperloop}, replaces the intermediate RGB round trip with a learned latent-space shortcut that approximates the same transformation while avoiding repeated VAE calls for decoding and encoding. Its second component, \textbf{Sparse Patch-based Decoding}, makes final reconstruction efficient by tracing patch seams to inconsistent GroupNorm statistics and reusing cached statistics across patches, enabling sparse VAE decoding with much smaller overlap. Integrated into the current state-of-the-art baseline, UltraFlash~reduces 4K generation latency by $\sim$ 2\(\times\) while improving image quality, demonstrating that transition-level optimization can advance both efficiency and fidelity. More broadly, these transition costs recur across megapixel synthesis pipelines, suggesting that optimizing pixel-space transitions is an effective route to fast, high-fidelity visual synthesis.

\end{abstract}

\section{Introduction}\label{sec:intro}

Latent diffusion models have greatly advanced visual synthesis, but their native training resolution remains a practical boundary. Although applications increasingly require 2K, 4K, and larger outputs, models trained around $512\times512$ or $1024\times1024$ often produce repeated objects, distorted layouts, or unstable textures when directly evaluated on larger canvases. Fine-tuning on high-resolution data is a natural remedy, but it is difficult to scale: diverse megapixel-scale datasets are limited, training at such resolutions is expensive, and the adapted model often becomes specialized to a fixed target resolution~\citep{xie2025sana,zhang2025diffusion}. Existing alternatives therefore rely on inference-time strategies that reuse pretrained models. Direct extrapolation methods modify inference on enlarged latents, but can suffer from memory cost and structural instability at extreme resolutions. Cascade refinement methods instead generate at a reliable native resolution and progressively construct larger images through refinement stages~\citep{du2024demofusion,huang2024fouriscale,qiu2025freescale,vontobel2025hiwave,zhang2025diffusion4k}. Representative systems differ in how they refine enlarged latents: DemoFusion uses progressive patch-based refinement, FouriScale and FreeScale modify frequency or scale behavior, HiWave uses structure-preserving inversion, and PixelRush~\citep{pixelrush} accelerates the cascade with one-step diffusion refinement. Among these, PixelRush currently provides the strongest speed--quality trade-off, making it the most relevant stress test for remaining non-denoising overhead. We therefore use PixelRush as the main instantiation of \methodname{}, while targeting VAE-heavy pixel-space transitions that are shared by the broader cascade family.

Most acceleration efforts for megapixel synthesis focus on denoising or refinement. This is natural, since diffusion sampling is expensive and larger spatial sizes increase the cost of each denoising operation. However, we find that this view is incomplete. Modern cascade pipelines also spend substantial time repeatedly moving representations through the VAE. We call these operations \emph{pixel-space transitions}: steps where a latent representation is converted to RGB or where RGB is converted back to a latent representation. These transitions appear in two places. First, intermediate latents are often decoded to RGB, resized, and re-encoded before refinement. Second, the final latent must be decoded to RGB, but at megapixel scale this is typically done patch by patch with dense overlap to suppress artifacts. As a result, a significant fraction of runtime is spent not on denoising itself, but on VAE-heavy transitions around denoising.

This observation exposes an underexplored source of overhead. The two transitions have different roles and therefore require different solutions. The intermediate RGB round trip is avoidable: it is used as a convenient way to transform a latent before refinement, but it need not be performed through pixels. The final transition is unavoidable because the output image must be RGB, but its dense patch implementation is inefficient. Thus, the design goal is to replace the avoidable intermediate VAE round trip and make the final VAE decoding sparse without introducing seams.

We present \textbf{\methodname}, which shifts the focus of acceleration from denoising alone to the VAE transitions around it. \methodname contains two components. First, \textbf{Latent Hyperloop} replaces the intermediate RGB round trip with a learned latent-space shortcut. It approximates the decode--resize--encode transformation while avoiding repeated VAE calls for decoding and encoding. Second, \textbf{Sparse Patch-based Decoding} accelerates final reconstruction. We trace sparse patch-decoding artifacts to inconsistent GroupNorm statistics across independently decoded patches. Sparse Patch-based Decoding fixes this by caching normalization statistics from reference patches and reusing them across all patches, enabling sparse VAE decoding with much smaller overlap.

Integrated into the current state-of-the-art cascade baseline \citep{pixelrush}, \methodname~ reduces 4K generation latency by $2\times$ while improving image quality. This result shows that VAE transitions remain a substantial optimization spot even in highly optimized megapixel synthesis systems. More broadly, RGB round trips and dense patch-wise VAE decoding appear across cascade pipelines, suggesting that optimizing pixel-space transitions is an effective route to faster, high-fidelity visual synthesis.

Our contributions are:
\begin{itemize}
    \item We identify repeated VAE-heavy pixel-space transitions as an underexplored source of overhead in megapixel visual synthesis, beyond denoising and refinement.
    \item We introduce \textbf{Latent Hyperloop}, a learned latent-space shortcut that replaces the intermediate decode--resize--encode round trip while avoiding repeated VAE decoding and encoding.
    \item We introduce \textbf{Sparse Patch-based Decoding}, which enables sparse VAE decoding by sharing cached normalization statistics across independently decoded patches.
\end{itemize}
\section{Related Work}
\label{sec:related}

\minisection{Megapixel Visual Synthesis.} Scaling diffusion models beyond their native training resolution has been explored through several routes. One route trains or adapts models for native high-resolution generation, such as SANA, HunyuanImage, and other dedicated 2K/4K generators~\citep{xie2025sana,HunyuanImage-2.1}. These methods can achieve strong quality, but they require high-resolution data, model adaptation, or dedicated training. A second route directly extrapolates pretrained models to larger canvases through inference-time modifications, such as convolutional dilation~\citep{he2023scalecrafter}, frequency or scale fusion~\citep{huang2024fouriscale}, flow-aligned guidance~\citep{bu2025hiflow}, or positional extrapolation for FLUX-based models~\citep{issachar2025dype}. These methods avoid retraining, but can suffer from memory cost, structural instability, or unnatural textures at extreme resolutions~\citep{lin2024accdiffusion}. A third route uses cascade or region-based refinement, synthesizing native-resolution windows and merging them into a larger image through average blending~\citep{bartal2023multidiffusion,du2024demofusion}, DDIM inversion~\citep{vontobel2025hiwave}, or few-step refinement~\citep{pixelrush}. This cascade line has progressed from multi-step patch refinement in DemoFusion to structure-preserving inversion in HiWave, and most recently to PixelRush, which uses one-step Turbo refinement to deliver the strongest speed--quality trade-off among  cascade methods.

Our work focuses on the cascade setting because it reuses pretrained models in their reliable operating regime while supporting megapixel outputs. We use PixelRush as the main experimental instantiation because it is already optimized for denoising speed; improving it therefore tests whether VAE transition overhead remains a meaningful bottleneck after the dominant sampling cost has been compressed. Rather than introducing another refinement strategy, \methodname{} optimizes the VAE-heavy transitions surrounding refinement: intermediate RGB round trips that prepare refinement inputs, and final patch-wise VAE decoding that reconstructs the output image.

\minisection{Pixel-Space Transitions.} Latent diffusion models use a VAE to move between compressed latents and RGB images~\citep{rombach2022high}. In megapixel synthesis, this creates two
transition costs. First, intermediate latents are often decoded to RGB, resized, and re-encoded before refinement. This route is stable and avoids
artifacts from naive latent manipulation~\citep{qiu2025freescale,vontobel2025hiwave}, but requires an expensive decode--encode round trip. Because the VAE encoder is nonlinear, ordinary latent interpolation is not that round trip. FAM Diffusion~\citep{yang2025fam} compensates with learned frequency and attention modulation over the whole latent, and it changes the denoiser. Latent Hyperloop is a patch-wise shortcut for the same decode--resize--encode transformation and leaves the denoiser unchanged.

Second, the final latent must be decoded to RGB. Since full-image VAE decoding is memory-intensive at megapixel scale, pipelines decode overlapping
patches and blend them. Large overlaps hide seams but repeat computation; small overlaps reveal boundary artifacts. We identify inconsistent GroupNorm
statistics~\citep{wu2018groupnorm} across patches as a key cause, and use \textbf{Sparse Patch-based Decoding} to share cached statistics across patches,
enabling sparse reconstruction with much smaller overlap.

\minisection{Super-Resolution.} Super-resolution methods reconstruct high-resolution images from low-resolution inputs~\citep{chen2024ntire,dong2014learning,wang2020deep}. Diffusion-based super-resolution methods~\citep{li2022srdiff,wu2024one} further improve perceptual quality, but are typically trained for restoration. Megapixel visual synthesis differs from restoration: the input is a generated intermediate representation, and the goal is coherent visual synthesis rather than recovering a unique high-resolution target~\citep{pixelrush}. Thus, although super-resolution and megapixel synthesis both involve scale expansion, they address different problem settings and require different evaluation criteria.
\section{Background}
\label{sec:background}

\minisection{Latent Diffusion Models~\citep{rombach2022high}} perform diffusion in the latent space of a pretrained autoencoder. Given an image \(\mathbf{x}\), the encoder \(\mathcal{E}\) maps it to a latent \(\mathbf{z}_0=\mathcal{E}(\mathbf{x})\). The forward process gradually corrupts this latent:
\begin{equation}
    q(\mathbf{z}_t \mid \mathbf{z}_0) =
    \mathcal{N}(\mathbf{z}_t;\sqrt{\bar{\alpha}_t}\mathbf{z}_0,(1-\bar{\alpha}_t)\mathbf{I}),
    \label{eq:forward}
\end{equation}
where \(t\) is the diffusion timestep and \(\bar{\alpha}_t\) is determined by the noise schedule. A denoising network \(\epsilon_\theta(\mathbf{z}_t,t)\) is trained to reverse this process. At inference, a sample is generated by denoising a Gaussian latent and decoding the final latent with the VAE decoder \(\mathcal{D}\).

\minisection{Cascade Refinement Pipelines.} Megapixel synthesis commonly starts from the model's native resolution and progressively enlarges the image through cascade
refinement~\citep{du2024demofusion,qiu2025freescale,vontobel2025hiwave,pixelrush}. Although methods differ in refinement details, each stage typically
initializes a larger latent by routing the current latent through RGB.

Let \(\mathbf{z}_0^r\) be a clean latent at resolution \(r\), where a \(2\times\) increase in height and width maps \(r\) to \(4r\). The standard pixel-space
transition is
\begin{equation}
  \bar{\mathbf{z}}_0^{4r}
  =
  \mathrm{UP}_{\mathrm{pixel}}(\mathbf{z}_0^r)
  =
  \mathcal{E}\!\left(
      \mathrm{Interpolate}(\mathcal{D}(\mathbf{z}_0^r),2)
  \right),
  \qquad
  \mathbf{z}_0^{4r} = \phi(\bar{\mathbf{z}}_0^{4r}),
  \label{eq:cascade_refinement}
\end{equation}
where \(\mathcal{D}\) and \(\mathcal{E}\) are the VAE decoder and encoder, \(\mathrm{Interpolate}(\cdot,2)\) resizes the RGB image by \(2\times\), and
\(\phi\) denotes refinement. This decode--resize--encode route is stable but VAE-heavy.

The final latent \(\mathbf{z}_0^R\) must also be decoded to RGB. Since full-image VAE decoding is memory-intensive at megapixel scale, practical systems
decode overlapping patches and stitch them together, where dense overlap suppresses seams but repeats computation. Thus, cascade pipelines contain two
major transition costs: repeated intermediate VAE round trips and final patch-wise VAE decoding. \methodname{} targets both: \textbf{Latent Hyperloop}
replaces the intermediate round trip with a learned latent shortcut, while \textbf{Sparse Patch-based Decoding} makes final reconstruction sparse and
consistent.
\section{Accelerating Megapixel Visual Synthesis}
\label{sec:method}
We present \textbf{\methodname}, which shifts acceleration from denoising alone to the VAE-heavy pixel-space transitions that surround it. In cascade megapixel synthesis, these transitions arise twice: intermediate latents undergo an RGB round trip before refinement, while final RGB decoding must be performed patch by patch due to memory constraints.

\begin{figure*}[t]
    \centering
    \includegraphics[width=1.0\textwidth]{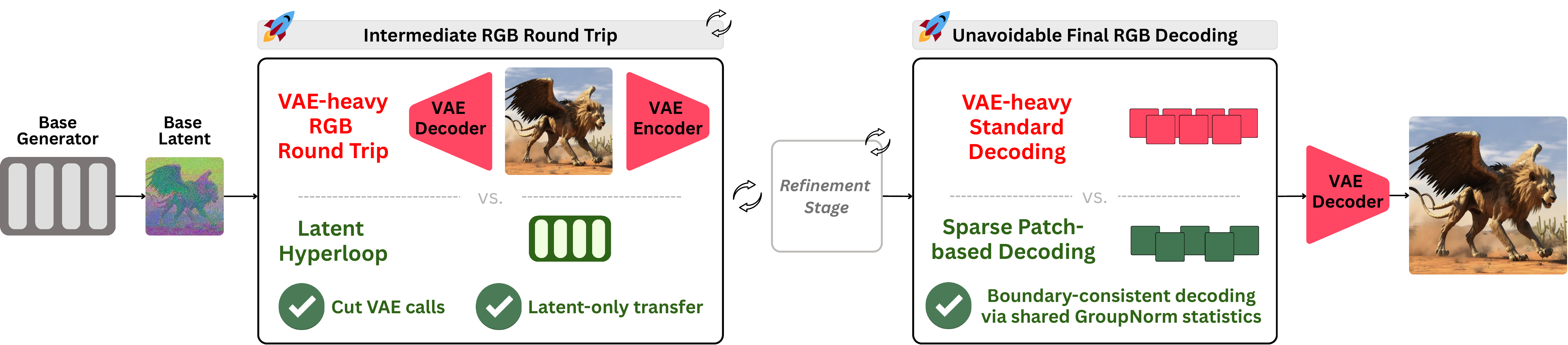}
    \caption{\textbf{\methodname{} targets transition costs common to cascade megapixel synthesis.}
    Cascade pipelines often use pixel space twice: an intermediate RGB round trip before refinement and dense patch-wise VAE decoding for final output. \methodname{} accelerates both without changing the denoising/refinement stage. \textbf{Latent Hyperloop} removes repeated VAE decode/encode calls during intermediate transfer, while \textbf{Sparse Patch-based Decoding} reduces final decoding redundancy by enforcing consistent GroupNorm statistics across patches.}    \label{fig:overview}
\end{figure*}

\methodname{} handles these two roles differently: \textbf{Latent Hyperloop} (Sec.~\ref{sec:latent_hyperloop}) replaces the avoidable intermediate RGB round trip, while \textbf{Sparse Patch-based Decoding} (Sec.~\ref{sec:cachevae}) accelerates the unavoidable final decoding. An overview is shown in Fig.~\ref{fig:overview}.

\subsection{Latent Hyperloop: Bypassing Intermediate RGB Round Trips}
\label{sec:latent_hyperloop}

\minisection{Motivation.} Many cascade refinement pipelines~\citep{du2024demofusion, qiu2025freescale, vontobel2025hiwave, pixelrush} initialize each refinement stage through a VAE-heavy RGB round trip. Given a latent \(\mathbf{z}_0^r\), the pipeline decodes it to RGB, changes its spatial scale, and re-encodes the resized image back into latent space:
\begin{equation}
    \mathrm{UP}_{\mathrm{pixel}}(\mathbf{z}_0^r)
    =
    \mathcal{E}\!\left(
    \mathrm{Interpolate}(\mathcal{D}(\mathbf{z}_0^r), 2)
    \right),
    \label{eq:rgb_transfer}
\end{equation}
where \(\mathcal{D}\) and \(\mathcal{E}\) denote the VAE decoder and encoder, respectively. The appeal of this RGB round trip is that resizing in pixel space is simple and well-defined. Its drawback is computational cost: each transfer invokes both the VAE decoder and encoder. In our main instantiation with the state-of-the-art cascade method \citep{pixelrush}, this transition accounts for approximately \textbf{\textit{30--40\%}} of cascade-stage runtime.

The challenge is that the RGB round trip cannot be replaced by ordinary latent interpolation. Since the VAE encoder is nonlinear, spatial resizing in latent space is not equivalent to resizing in RGB and then encoding the result: $
    \mathrm{Interpolate}(\mathbf{z},2)
    \neq
    \mathcal{E}\!\left(
    \mathrm{Interpolate}(\mathcal{D}(\mathbf{z}),2)
    \right)$.
Thus, we replace generic latent interpolation with a learned latent-space shortcut that approximates the decode--resize--encode transformation while avoiding the VAE decode--encode computation.

\minisection{Problem Setting.}
We define \textbf{Latent Hyperloop} as a lightweight function \(\mathcal{H}_{\phi}\) that maps the current latent directly to the latent that would otherwise be produced by the RGB round trip:
\begin{equation} \bar{\mathbf{z}}_0^{4r} = \mathcal{H}_{\phi}(\mathbf{z}_0^r) \approx \mathrm{UP}_{\mathrm{pixel}}(\mathbf{z}_0^r). \label{eq:latent_hyperloop} \end{equation}

Once trained, \(\mathcal{H}_{\phi}\) replaces the decode--interpolate--encode route during intermediate transfer. A good Latent Hyperloop \(\mathcal{H}_\phi\) must satisfy the following conditions: (1) \textbf{\textit{Accurate}} in replicating the upsampling operator $\mathrm{UP}_{\mathrm{pixel}}$ without going through pixel space, (2) \textbf{\textit{Efficient}} in both memory usage and computation, causing minimal latency, and (3) \textbf{\textit{General}} to work at every cascade step. These criteria are challenging, requiring novel designs in data flow, network structure, training data, and training objectives. We will discuss below our proposals to handle each mentioned aspect.

\minisection{Context-Expanded Patching.} To make $\mathcal{H}_\phi$ general and work at every cascade step regardless of input resolution, we design it to process patch-by-patch, using fixed resolutions for the input and output latent patches. Since the follow-up refinement step also processes overlapping patches, a convenient design is to use the same patch configuration in these two steps: the output patches of $\mathcal{H}_\phi$ should have the same resolution and location as the ones used in the refinement. With that configuration, the patches produced by $\mathcal{H}_\phi$ can be sent for processing at the refinement step directly, skipping the operation to collect and aggregate them to form the full latent output $\bar{\mathbf{z}}_0^{4r}$.

For each target patch in \(\bar{\mathbf{z}}_0^{4r}\), a natural input to \(\mathcal{H}_{\phi}\) is the coordinate-matched source patch in \(\mathbf{z}_0^r\). However, this crop alone is insufficient. The decode--resize--encode route approximated by \(\mathcal{H}_{\phi}\) contains convolutional VAE layers, so the target patch depends not only on the aligned source region but also on its surrounding context. We therefore expand each source crop by \(p\) context pixels on every side before feeding it to \(\mathcal{H}_{\phi}\), so that a source patch of side length \(s\) becomes an input of side length \(s+2p\); we use \(s=64\) and \(p=8\), i.e., \(80\times80\) inputs for \(128\times128\) target patches. This context-expanded design captures the local dependencies of the original RGB round trip while preserving patch-wise processing. Its necessity is confirmed by the ablation on \(p\) in Appendix~\ref{app:context_ablation}.

\minisection{Lightweight Spatial Prior.}
\label{sec:bridge_arch}
Latent Hyperloop \(\mathcal{H}_\phi\) needs to double the resolution of the input latent $\mathbf{z}_0^{r}$. While we could use any super-resolution backbone with an adjusted number of input/output channels and train on paired data, this would require the model to learn spatial upsampling behavior in the latent space from scratch. Instead, we reuse \textit{the first decoding block} of the VAE decoder, denoted as $\mathcal{D}^{(1)}$, which already implements latent upsampling and has been well trained on massive data. This pretrained block serves as a \textit{spatial prior}, effectively guiding $\mathcal{H}_\phi$'s spatial upsampling operation. However, this block often produces feature maps with a different number of channels and unaligned geometry compared to the desired output $\mathbf{z}_0^{4r}$. Hence, we need \textit{a projection head} to adjust the number of channels and refine the geometry of the produced latent output. 

In summary, our proposed \textbf{Latent Hyperloop} \(\mathcal{H}_\phi\) consists of two components: \textbf{\textit{a frozen VAE decoding block}} \(\mathcal{D}^{(1)}\) and \textbf{\textit{a trainable projection head}} \(h_\psi\). Because context-expanded patching uses an input crop larger than the coordinate-matched source region, the spatial size entering \(\mathcal{H}_\phi\) is larger than the nominal source patch size. We therefore adjust the padding in \(\mathcal{D}^{(1)}\) and the early layers of \(h_\psi\) so that the output matches the required target patch size, requiring no major architectural change.

\minisection{Training Data.}
\label{sec:bridge_data}
Training the \textbf{Latent Hyperloop} requires paired input/output data. Given an image $\mathbf{x}$ at the base resolution $r_0$, e.g., $1024\times1024$, and its upsampled version $\mathbf{x}^{up}$, we  pass them through the VAE encoder to get the base latent $\mathbf{z}_0^{r_0} = \mathcal{E}(\mathbf{x})$ and its upsampled latent $\bar{\mathbf{z}}_0^{4r_0} = \mathcal{E}(\mathbf{x}^{up})$. We then crop these latents with predefined cropping boxes, denoted as $b_i$ for $\mathbf{z}_0^{r_0}$ and $b^{up}_i$ for $\bar{\mathbf{z}}_0^{4r_0}$ with $i$ being the patch index among predefined patch locations, to form paired input/output latent patches $\left(\mathbf{z}_0^{r_0}(b_i), \bar{\mathbf{z}}_0^{4r_0}(b^{up}_i)\right)$. We construct these pairs from a dataset of high-resolution high-quality images. 

When the VAE is perfect, i.e., $\mathcal{D}(\mathcal{E}(\mathbf{x})) = \mathbf{x}$, and $\mathbf{x}^{up} = \mathrm{Interpolate}(\mathbf{x}, 2)$, the upsampled latent $\bar{\mathbf{z}}_0^{4r_0}$ is equal to the result of applying VAE round-trip on the base latent $\mathbf{z}_0^{r_0}$, i.e., $\bar{\mathbf{z}}_0^{4r_0} = \mathrm{UP}_{\mathrm{pixel}}(\mathbf{z}_0^{r_0})$. However, in practice, VAE is imperfect, leading to imprecise upsampling when using the VAE round-trip. Our data construction avoids that flaw, allowing \methodname{} to even surpass the original cascade baseline with VAE round trip in terms of quality, as will be shown in \Cref{sec:experiments}.

\minisection{Training Objectives.}
\label{sec:bridge_objectives}
After constructing training data, we define training losses for building a high-quality Latent Hyperloop $\mathcal{H}_\phi$. Given each training example $(\mathbf{z}^{in}, \mathbf{z}^{out})$, two complementary losses are applied to ensure that $\mathcal{H}_\phi$ could faithfully estimate $\mathbf{z}^{out}$ from $\mathbf{z}^{in}$: $\hat{\mathbf{z}}^{out} = \mathcal{H}_\phi(\mathbf{z}^{in}) \approx \mathbf{z}^{out}$.

\myparagraph{Latent Reconstruction ($\Loss_{rec}$).}
A straightforward objective is latent reconstruction, implemented by an $\ell_1$ reconstruction loss:
\begin{equation}
    \Loss_{rec} = \| \mathbf{z}^{out} - \hat{\mathbf{z}}^{out} \|_1.
\end{equation}
While $\ell_2$ is a common choice, it is less effective for latent-space reconstruction, as noted in recent works \citep{sdxl_lightning, hipa, improved_shortcut}. The $\ell_1$ objective provides better training stability and preserves the structural sparsity of the upsampled latent.

\myparagraph{Attention Alignment ($\Loss_{atten}$).}
Reconstruction loss alone is insufficient.
Since the loss is operated directly on the latent manifold, a small numerical distance error does not necessarily guarantee that two latents are perceptually or semantically equivalent. Furthermore, by treating each latent pixel independently, a simple reconstruction objective often tolerates subtle misalignments. Both lead to a loss of fine-grained structure in the decoded image.
In principle, a GAN loss could help, but it often makes optimization substantially more sensitive to hyperparameters and empirically less stable.

Motivated by X2I \citep{X2I}, we instead seek supervision signals from attention responses of the refiner model when processing $\mathbf{z}^{out}$ and $\hat{\mathbf{z}}^{out}$, which are stable and sensitive to fine structure. Self-attention maps in the refiner encode rich interactions across spatial locations in the latent, and thus can \emph{magnify} small deviations into larger, more detectable discrepancies in the network’s internal responses. By monitoring the attention misalignment, we can penalize structured mismatches between $\mathbf{z}^{out}$ and $\hat{\mathbf{z}}^{out}$ that $\Loss_{rec}$ may underweight.

Let $\epsilon_\theta(\cdot,t,c)$ denote the refiner's denoising function at time step $t$ with text conditioning $c$. 
To compare its responses under identical diffusion conditions, we apply the forward noising operator using the \emph{same} noise drawn from $\mathcal{N}(0,I)$ to convert $\mathbf{z}^{out}$ and $\hat{\mathbf{z}}^{out}$ into noised latents $\mathbf{z}_t^{out}$ and $\hat{\mathbf{z}}_t^{out}$. We then run one denoising pass with $\epsilon_\theta$ on these noised inputs and extract the self-attention map $P^l(\cdot;t,c)$ for each self-attention (SA) layer $l$. 
We minimize the reverse KL divergence between the attention distributions induced by $\mathbf{z}_t^{out}$ and $\hat{\mathbf{z}}_t^{out}$:
\begin{equation}
\Loss_{atten}
=\sum_{l\in\text{SA layers}} \tau^2\,
\mathrm{KL}\!\left(
P^l(\mathbf{z}_t^{out};t,c)
\,\middle\|\,
P^l(\hat{\mathbf{z}}_t^{out};t,c)
\right),
\end{equation}
where $\tau$ is a temperature used for numerical stability. Empirically, we found this attention alignment exceptionally effective, speeding up training convergence and enhancing fine-detail reconstruction.

\myparagraph{Final Objective ($\Loss_{final}$).}
Our training minimizes a weighted sum of latent reconstruction and attention alignment, using hyperparameters $\alpha,\beta\ge 0$:
\begin{equation}
\Loss_{final}=\alpha \Loss_{rec}+\beta \Loss_{atten}.
\label{eq:bridge_loss}
\end{equation}
For simplicity, we set $\alpha=\beta=1$ in our experiments.
\subsection{Sparse Patch-based Decoding: Accelerating Final RGB Decoding}
\label{sec:cachevae}

\begin{figure}[t]
\centering
\includegraphics[width=\columnwidth]{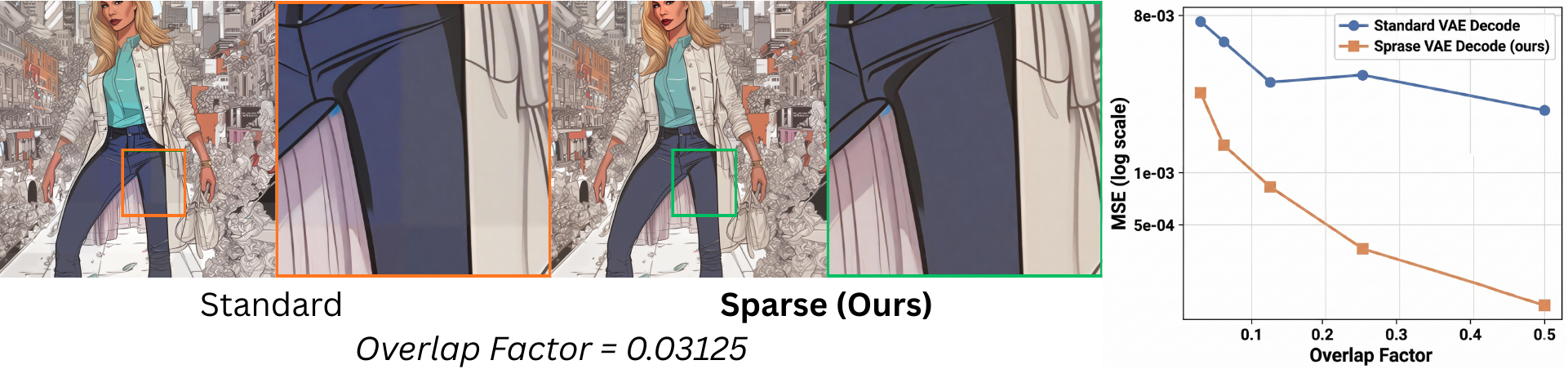}
\caption{\textbf{Shared GroupNorm statistics enable sparse final RGB decoding.} When overlap is small, standard patch-wise decoding shows seam artifacts caused by independently computed GroupNorm statistics. \textbf{Sparse Patch-based Decoding} uses shared cached statistics to align normalization across patches, making adjacent patches decode consistently. The right plot shows that our method maintains consistently low boundary MSE across overlap factors when decoding with sparse overlaps.}
\label{fig:groupnorm_artifact}
\end{figure}

While Latent Hyperloop removes avoidable intermediate RGB round trips, the final pixel-space transition is unavoidable: after refinement, the latent \(\mathbf{z}_0^R\) must be decoded into RGB. At megapixel scale, however, full-latent VAE decoding is often memory-prohibitive, so practical synthesis pipelines decode the latent patch by patch and blend overlapping regions~\citep{bartal2023multidiffusion, du2024demofusion, vontobel2025hiwave, pixelrush}. Dense overlap helps suppress seams, but it also repeats VAE computation over many regions. A natural acceleration is to reduce the overlap, but naive sparse decoding introduces visible boundary artifacts, including color shifts, brightness inconsistencies, and texture discontinuities at patch boundaries, as shown in Fig.~\ref{fig:groupnorm_artifact}.  

\minisection{Why Sparse Decoding Fails.} We trace the root cause of sparse-decoding artifacts to the \textit{GroupNorm (GN) layers} in the VAE decoder. GroupNorm~\citep{wu2018groupnorm} normalizes activations within each group of channels using statistics computed from the current input. 

When the full image is decoded in a single pass, the mean \(\mu\) and variance \(\sigma^2\) are computed over the \emph{entire} spatial extent, producing globally consistent normalization. However, when decoding individual patches, each patch computes its own \(\mu^{(j)}\) and \(\sigma^{2(j)}\) based only on its local spatial content. 

Patches with different spatial content (\eg, a bright sky vs.\ a dark ground) yield substantially different statistics. This causes the same latent values in overlapping regions to be decoded to different pixel values depending on which patch they belong to, producing visible seams even after blending.

\minisection{Shared GN Statistics.} 
Our solution is simple and requires no training. We compute a set of \textit{reference GroupNorm statistics} once, then reuse them for all patches during decoding. First, we decode a set of reference patches $\{P_{\mathrm{ref}}^{(i)}\}_{i=1}^{N_{\mathrm{ref}}}$ (utilizing the four corner patches of the latent) through the VAE decoder, recording the GroupNorm statistics $(\mu_l^{\mathrm{ref}}, \sigma_l^\mathrm{ref})$ at every layer $l$.
The cached statistics are $\mathcal{C} = \{(\mu_l^{\mathrm{ref}}, \sigma_l^\mathrm{ref})\}_{l=1}^{L}$,
where $L$ is the number of GroupNorm layers in the VAE decoder.
Second, when decoding each patch $P^{(j)}$, we replace the per-patch GroupNorm computation with the cached reference statistics.
This ensures that identical latent values always produce identical pixel values, regardless of which patch they belong to. The overlapping regions now agree perfectly, and simple blending, or even direct stitching, produces seamless results.

\minisection{Enabling Smaller Overlap.} A key benefit of Sparse Patch-based Decoding is that it enables the use of much smaller overlaps during patch-wise VAE decoding. In standard cascade and patch-based synthesis pipelines~\citep{bartal2023multidiffusion, du2024demofusion, vontobel2025hiwave, pixelrush}, large overlaps (e.g., 25\%) are often needed to mask patch-boundary inconsistencies through blending. With cached GroupNorm statistics, the overlap can be reduced by up to \textbf{8\(\times\)} in our main setting while preserving visual continuity, since it no longer needs to hide per-patch normalization mismatch. This directly translates to fewer total patches and faster decoding:
\begin{equation}
N_{\mathrm{patches}} = \left\lceil \frac{H - o}{H_p - o} \right\rceil \times \left\lceil \frac{W - o}{W_p - o} \right\rceil,
\label{eq:num_patches}
\end{equation}
where \(H_p \times W_p\) is the patch size and \(o\) is the overlap. Smaller overlap increases the patch stride and reduces \(N_{\mathrm{patches}}\); in our main 4K setting, this reduces the number of decoded patches by up to \textbf{30--40\%}, leading to a proportional reduction in VAE decoding time. Fig.~\ref{fig:groupnorm_artifact} quantifies boundary consistency across overlap settings. We vary the overlap factor and measure the MSE between decoded pixel values in overlapping regions of adjacent patches, averaged over 100 held-out latents. Standard decoding exhibits high boundary error that decreases only gradually with larger overlaps, while Sparse Patch-based Decoding maintains low boundary error by enforcing globally consistent GroupNorm statistics.

\section{Experiments}
\label{sec:experiments}

\subsection{Experimental Settings}
\noindent\textbf{Latent Hyperloop.}
Latent Hyperloop is a lightweight network with $\sim5$M trainable parameters. Unless otherwise stated, we use SwinIR~\citep{liang2021swinir} as the projection head backbone and train it with the objectives described in Sec.~\ref{sec:bridge_objectives}. For the attention-alignment loss, we use a fixed noising level \(t=1\) and temperature \(\tau=3.0\). Training data consists of 1.2M images synthesized with SDXL~\citep{podell2023sdxl} using prompts from JourneyDB~\citep{pan2023journeydb} and LAION~\citep{schuhmann2022laion}, each paired with its \(2\times\)-interpolated version as described in Sec.~\ref{sec:bridge_data}. Training uses 8 NVIDIA H100 GPUs (80GB each) with a batch size of 32 per GPU, i.e., a global batch size of 256; optimization details are given in Appendix~\ref{app:lh_details}.

\noindent\textbf{Evaluation Data and Metrics.}
Following prior protocols~\citep{du2024demofusion, huang2024fouriscale, qiu2025freescale}, we evaluate on 1000 prompts sampled from LAION-Aesthetics~\citep{schuhmann2022laion,laion2022aesthetics} and report Fr\'{e}chet Inception Distance (FID)~\citep{heusel2017gans}, crop-based FID (FID$_c$), Inception Score (IS)~\citep{salimans2016improved}, and wall-clock inference time. FID measures global distribution quality, while FID$_c$ evaluates local high-frequency fidelity using crops without resizing. All latency numbers are measured on a single NVIDIA A100-40GB GPU unless stated otherwise.

\noindent\textbf{Baselines.} Our main comparison focuses on SDXL-based cascade methods, including FouriScale~\citep{huang2024fouriscale}, DemoFusion~\citep{du2024demofusion}, FreeScale~\citep{qiu2025freescale}, and the fast cascade baseline \basename{}~\citep{pixelrush}. This setting provides a direct comparison because these methods share a similar base-model regime and target high-resolution extrapolation from pretrained native-resolution generators. We use \basename{} as the representative fast cascade pipeline because it offers the strongest speed--quality trade-off among prior cascade methods. Unlike FouriScale, DemoFusion, and FreeScale, which rely on multi-step SDXL refinement, \basename{} uses one-step Turbo refinement for rapid generation. We denote configurations by the refiner used in the cascade: \textbf{-SD} uses SD-Turbo, while \textbf{-XL} uses SDXL-Turbo~\citep{sauer2024adversarial}. Unless otherwise specified, both configurations follow the same base-generation protocol. Finally, \methodname{} denotes the \basename{} fast cascade equipped with our Latent Hyperloop and Sparse Patch-based Decoding.

\noindent\textbf{Refinement-stage sparse patching.}
\basename{} already uses sparse patching during diffusion refinement to reduce overlap between refinement patches, discussed in its supplemental materials. We regard this as part of the baseline cascade pipeline.
It is separate from \textbf{Sparse Patch-based Decoding}: the former reduces denoising/refinement cost, while the latter reduces final VAE-to-RGB decoding cost by enabling small-overlap patch decoding without boundary artifacts. For fair comparison, we report \basename{} under the same refinement-stage sparse patching setting.

\subsection{Main Results on Fast Cascade Inference}

\begin{table*}[t]
    \centering
    \scriptsize
    \caption{\textbf{Quantitative comparison among SDXL-based high-resolution generation methods.} We report FID ($\downarrow$), FID$_c$ ($\downarrow$), Inception Score (IS $\uparrow$), and Inference Time (seconds).}
    \label{tab:main_results}
    \resizebox{\textwidth}{!}{
    \begin{tabular}{l cccc cccc}
        \toprule
        & \multicolumn{4}{c}{\textbf{Resolution 2K} ($2048 \times 2048$)} & \multicolumn{4}{c}{\textbf{Resolution 4K} ($4096 \times 4096$)} \\
        \cmidrule(lr){2-5} \cmidrule(lr){6-9}
        \textbf{Method} & \textbf{FID} $\downarrow$ & \textbf{FID$_c$} $\downarrow$ & \textbf{IS} $\uparrow$ & \textbf{Time (s)} & \textbf{FID} $\downarrow$ & \textbf{FID$_c$} $\downarrow$ & \textbf{IS} $\uparrow$ & \textbf{Time (s)} \\
        \midrule
        FouriScale~\citep{huang2024fouriscale}    
            & 72.65 & 51.90 & 12.31 & 87.84  
            & 98.97 & 69.56 & 8.54  & 680.57 \\
        DemoFusion~\citep{du2024demofusion}       
            & 68.46 & 43.39 & 13.15 & 75.89  
            & 74.75 & 52.14 & 12.57 & 507.74 \\
        FreeScale~\citep{qiu2025freescale}        
            & 52.87 & 29.64 & 13.56 & 53.20  
            & 58.28 & 34.01 & 13.35 & 323.93 \\
        \midrule
        \basename{}-SD~\citep{pixelrush}         
            & 52.56 & 29.66 & 13.69 & \underline{3.17}   
            & 56.50 & 33.76 & 13.42 & 16.37  \\
        \basename{}-XL~\citep{pixelrush}         
            &50.13  &\textbf{29.13}  &\textbf{14.32}  &4.06    
            & 54.28 & 33.15 & 13.71 & 17.11  \\
        \midrule
        \rowcolor{green!8}
        \textbf{\methodname{}-SD} (Ours)        
            & \underline{50.02} & \underline{29.21} & 14.03 & \textbf{2.78} 
            & \underline{52.38} & \textbf{32.47} & \textbf{13.99} & \textbf{8.04} \\
        \rowcolor{green!8}
        \textbf{\methodname{}-XL} (Ours)        
            & \textbf{49.86} & 29.73 & \underline{14.14} & 3.50
            & \textbf{51.32} & \underline{33.01} & \underline{13.93} & \underline{9.61} \\
        \bottomrule
    \end{tabular}}
\end{table*}

\begin{figure*}[t]
    \centering
\includegraphics[width=\textwidth]{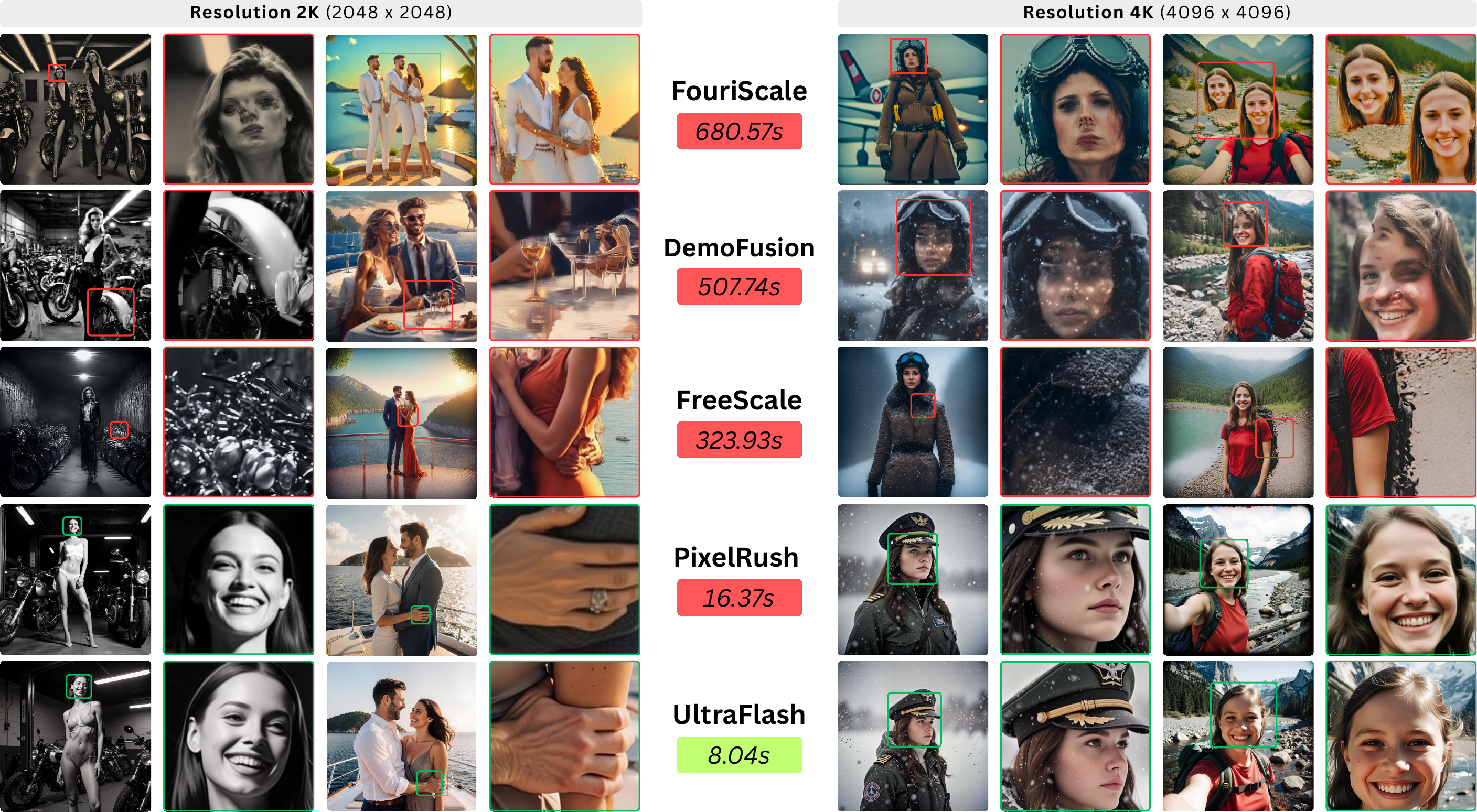}
\caption{\textbf{Qualitative comparison at 2K and 4K resolution.}
    We compare high-resolution generation from FouriScale, DemoFusion, FreeScale, PixelRush, and UltraFlash at $2048\times2048$ (left) and $4096\times4096$ (right). UltraFlash produces sharper textures, cleaner facial regions, and more coherent fine structures, while achieving substantially faster inference. Runtimes are measured at $4096\times4096$.}
    \label{fig:qualitative}
\end{figure*}

Table~\ref{tab:main_results} shows that adding our transition operators to the fast cascade improves both efficiency and fidelity. At 2K, the transition overhead is relatively small because the cascade contains fewer high-resolution patches, yet the gains remain consistent: \methodname{}-SD reduces runtime from 3.17s to 2.78s while improving FID from 52.56 to 50.02, and \methodname{}-XL achieves the best 2K FID of 49.86. The benefit becomes more pronounced at 4K, where transition costs compound across cascade stages and final VAE decoding. \methodname{}-SD reduces latency from 16.37s to 8.04s relative to the corresponding \basename{}-SD setting, while improving FID from 56.50 to 52.38. The XL configuration shows the same trend, reducing latency from 17.11s to 9.61s and improving FID from 54.28 to 51.32. These results indicate that replacing the intermediate RGB round trip with \textbf{Latent Hyperloop} and accelerating final RGB decoding with \textbf{Sparse Patch-based Decoding} removes transition overhead without sacrificing visual quality. Detailed ablations of both components are provided in Appendix~\ref{app:quantitative}.

\subsection{Qualitative Comparison}
As illustrated in Fig.~\ref{fig:qualitative}, \methodname{}    demonstrates superior structural coherence and realism compared to existing state-of-the-art methods. While baselines like FouriScale and DemoFusion exhibit structural deterioration at high resolutions (e.g., object repetition or unnatural texture), both \basename{} and \methodname{} maintain robust global layout. 
However, the advantage of \methodname{} becomes evident in the fine-grained details. As shown in the zoomed crops, our pipeline resolves sharper high-frequency textures, effectively mitigating the subtle smoothing artifacts observed in \basename{}.

\subsection{Comparison with Non-Cascade 4K Routes}
\label{sec:noncascade_4k}
To position our cascade pipeline in the broader 4K generation landscape, Table~\ref{tab:additional_results}(a) compares against two representative non-cascade routes. SANA-4K~\citep{xie2025sana} directly targets native 4K generation with a compact high-resolution model design, so it serves as a speed-oriented native-generation baseline. DyPE-FLUX~\citep{issachar2025dype} extrapolates a strong FLUX~\citep{flux2024} backbone to 4K and serves as a quality-oriented direct-inference baseline. Because these methods differ in backbone, training data, and inference assumptions, we treat this as a contextual comparison. 
\methodname{}-SD outperforms both methods in FID and particularly in latency, suggesting that a fast cascade pipeline with optimized pixel-space transitions can remain highly competitive with native or direct 4K approaches. We exclude super-resolution post-processing baselines from this comparison since they address a different restoration-style setting and have been shown to underperform cascade generation~\citep{qiu2025freescale}.

\begin{table*}[t]
\centering
\caption{\textbf{Additional evaluations beyond the main SDXL cascade comparison.} Left: contextual comparison with alternative 4K generation routes, separated from Table~\ref{tab:main_results} because these methods use different backbones or training assumptions. Right: transfer of the proposed transition operators to a second high-resolution pipeline.}
\begin{minipage}[t]{0.47\textwidth}
\centering
\small
\resizebox{\textwidth}{!}{%
\begin{tabular}{lcccc}
\toprule
Method & Backbone & Route & FID$\downarrow$ & Time (s)$\downarrow$ \\
\midrule
SANA-4K~\citep{xie2025sana} & SANA & Native 4K & 72.13 & 24.79 \\
DyPE-FLUX~\citep{issachar2025dype} & FLUX & Direct 4K & 59.05 & 587.13 \\
\midrule
\methodname{}-SD$^*$ & SDXL/SD-Turbo & Cascade & 52.38 & 8.04 \\
\methodname{}-XL$^*$ & SDXL/SDXL-Turbo & Cascade & 51.32 & 9.61 \\
\bottomrule
\end{tabular}}
\textbf{(a) Alternative 4K generation routes}
\end{minipage}
\hfill
\begin{minipage}[t]{0.50\textwidth}
\centering
\small
\resizebox{\textwidth}{!}{%
\begin{tabular}{lcccc}
\toprule
Configuration & FID$\downarrow$ & FID$_c$ $\downarrow$ & IS$\uparrow$ & Time (s)$\downarrow$ \\
\midrule
FreeScale~\citep{qiu2025freescale} & 58.28 & 34.01 & 13.35 & 323.93 \\
+ \methodname{} & \textit{54.42} & \textit{34.58} & \textit{13.66} & \textit{313.40} \\
\bottomrule
\end{tabular}}
\textbf{(b) Cross-pipeline generalization}
\end{minipage}
\label{tab:additional_results}
\end{table*}

\subsection{Plug-in Transfer}
\label{sec:plugin_transfer}

To test whether \methodname{} is more than a \basename{}-specific optimization, we apply the same transition operators to
FreeScale~\citep{qiu2025freescale}. As shown in Table~\ref{tab:additional_results}(b), this improves FID from 58.28 to 54.42 and IS from 13.35 to 13.66.
Runtime changes modestly from 323.93s to 313.40s because FreeScale remains dominated by multi-step denoising. This confirms that the operators are
reusable, while the \basename{} results show their full speed benefit when VAE transitions become the main remaining bottleneck.

\subsection{User Study}
\label{sec:user_study}

\begin{figure}[t]
    \centering
    \includegraphics[width=\linewidth]{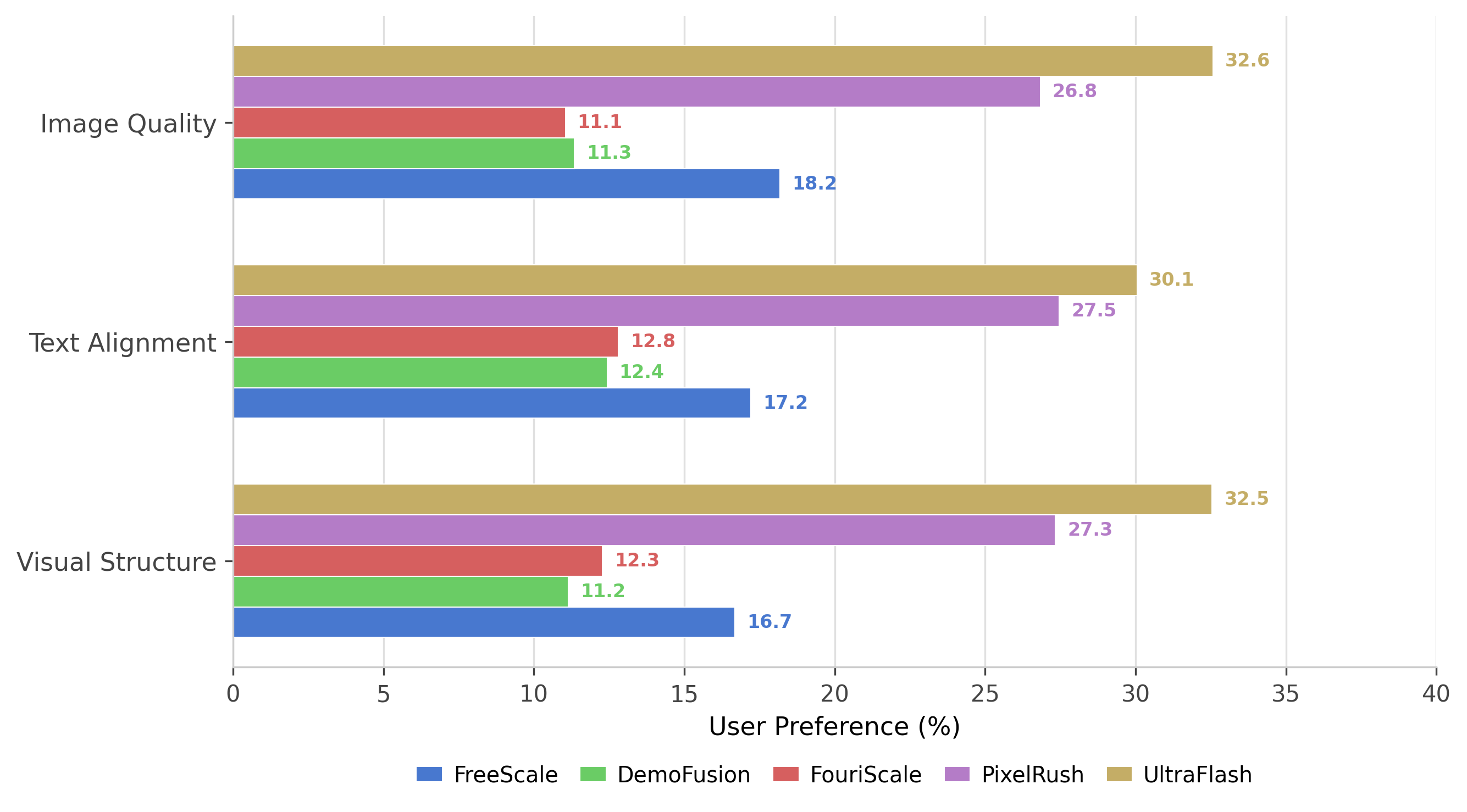}
    \caption{\textbf{User preference study.} Preference share of each method on three criteria. \methodname{} (gold bars) receives the highest share ($\sim$30--33\%) on all three.}
    \label{fig:user_study}
\end{figure}

To complement the automated metrics with human judgment, we conducted a user study with 137 participants on 25 diverse prompts (3,425 comparisons). Participants compared \methodname{} with the baselines on three criteria: \textbf{Text Alignment}, \textbf{Image Quality}, and \textbf{Visual Structure}. As shown in Fig.~\ref{fig:user_study}, \methodname{} receives the highest preference rate on all three criteria. This is consistent with the qualitative comparison in Fig.~\ref{fig:qualitative}, where \methodname{} renders sharp, high-frequency textures without the tiling artifacts and structural inconsistencies observed in competing methods.

\section{Conclusion}
\label{sec:conclusion}

We presented \methodname{}, a transition-efficient method for megapixel visual synthesis. Instead of only accelerating denoising, \methodname{} targets VAE-heavy pixel-space transitions in cascade pipelines. \textbf{Latent Hyperloop} replaces the intermediate decode--resize--encode round trip with a
learned latent shortcut, while \textbf{Sparse Patch-based Decoding} accelerates final RGB reconstruction by sharing cached GroupNorm statistics across
patches. Integrated into a state-of-the-art cascade baseline, \methodname{} reduces 4K latency by \(2\times\) while improving image quality, showing that
pixel-space transitions are an effective target for improving the speed--quality trade-off of megapixel synthesis.

\begin{ack}
This work was funded by Qualcomm Technologies, Inc. The authors declare no competing interests.
\end{ack}

\bibliographystyle{abbrv}
\bibliography{main}

\clearpage
\appendix

\clearpage
\setcounter{section}{0}
\setcounter{figure}{0}
\setcounter{table}{0}
\setcounter{equation}{0}
\setcounter{algorithm}{0}

\renewcommand{\thesection}{\Alph{section}}
\renewcommand{\thefigure}{A\arabic{figure}}
\renewcommand{\thetable}{A\arabic{table}}
\renewcommand{\theequation}{A\arabic{equation}}
\renewcommand{\thealgorithm}{A\arabic{algorithm}}
\title{\methodname{}: Eliminating the VAE Bottleneck for Ultra-Fast High-Resolution Image Generation} 
\maketitle
\begin{center}
    \Large\textbf{Supplementary Material}
\end{center}

\vspace{0.5cm}

\section{\methodname{} Pipeline}
\label{app:implementation}
\subsection{Implementation Details}
\noindent\textbf{Base Generation.}
Following previous works, we use SDXL as the default base model to generate images at its native $1024\times1024$ resolution with a classifier-free guidance scale of 7.5. The reverse diffusion process uses deterministic DDIM sampling ($\eta = 0$).

\noindent\textbf{Cascade Upsampling.}
Our pipeline operates in a cascaded manner, doubling the spatial resolution at each level. For example, generating an $8192\times8192$ image proceeds as $1024 \to 2048 \to 4096 \to 8192$, with refinement applied at each level. At each cascade level, the Latent Hyperloop adapter upsamples the latent by $2\times$. Since the adapter backbone is lightweight ($\sim$4.8M parameters), the computational overhead of processing overlapping regions is negligible, so we use a generous 50\% overlap for the upsampling stage to ensure smooth transitions.

\noindent\textbf{Patch Extraction.}
For the refinement stage, we support two patching strategies. In \emph{dense patching}, the upsampled latent is tiled into patches with 50\% overlap along both spatial dimensions. In \emph{sparse patching}, we reduce the total number of patches for faster inference. Given the target number of patches per row (or column) as $e$, the fixed patch window size $w$, and the upsampling factor relative to the original patch window size $u$, we need to compute the overlap size for interior patches $o$ and the last patch $o_{\mathrm{last}}$ that satisfy:
\begin{equation}
    o \cdot (e - 2) + o_{\mathrm{last}} = (e - u) \cdot w
    \label{eq:sparse_overlap}
\end{equation}
To distribute the overlaps almost evenly, we compute them as follows:
\begin{align}
    o & = \lfloor \frac{(e - u) \cdot w}{e - 1} \rfloor,\\
    o_{\mathrm{last}} & = (e - u) \cdot w - o \cdot (e - 2)
    \label{eq:sparse_overlap_compute}
\end{align}
 Table~\ref{tab:patch_config} summarizes the patching configurations. Note that the patch size for SD-Turbo ($64 \times 64$) is only half of the patch size for SDXL-Turbo ($128 \times 128$); hence, for the same target image resolution, SD-Turbo requires a doubled upsampling scale $u$. At $4\times$ the window size (e.g., SDXL-Turbo at 4K or SD-Turbo at 2K), sparse patching uses 6 patches per row/column instead of the 7 required by dense 50\% overlap. At $8\times$ the window size (e.g., SD-Turbo at 4K), the reduction is more substantial: 11 patches per row/column versus 15 with dense patching. Note that SDXL-Turbo at 2K cannot benefit from sparse patching, as dense patching already produces only 3 patches along each dimension, which is the minimum needed for full coverage. Combined with sparse patching, SD-Turbo configurations achieve the best speed-quality trade-off in our experiments.

\setlength{\tabcolsep}{4pt}
\begin{table}[h]
    \centering
    \small
    \caption{\textbf{Patching configurations.} We report here the patch count per spatial dimension (total patches = count$^2$) in each setting.}
    \label{tab:patch_config}
    \begin{tabular}{lccc}
        \toprule
        \textbf{Setting} & \textbf{Scale} & \textbf{Dense} & \textbf{Sparse} \\
        \midrule
        SDXL-Turbo, 2K & $\times$4 & 3 & 3 \\
        SDXL-Turbo, 4K / SD-Turbo, 2K & $\times$4 & 7 & 6 \\
        SD-Turbo, 4K & $\times$8 & 15 & 11 \\
        \bottomrule
    \end{tabular}
\end{table}

\subsection{Pseudo Algorithm}
Integrating both components, the complete \methodname{} pipeline is summarized in Algorithm~\ref{alg:\methodname{}}. Note that the VAE is invoked only \textbf{once} at the very end (for final decoding), compared to $2\times s$ times in \basename{} (encode + decode at each of $s$ cascade levels). The Sparse VAE decoding further accelerates this final decode by enabling small overlaps.

\begin{algorithm}[t]
\caption{\methodname{} Pipeline}
\label{alg:\methodname{}}
\begin{algorithmic}[1]
\Require Base latent $\mathbf{z}_0^n$ at resolution $n$, target $R = n \times 4^s$
\Require Inversion timestep $K$, noise injection weight $\lambda$
\Require LH adapter $\mathcal{H}_\theta$, few-step denoiser $\epsilon_\gamma^{\mathrm{refine}}$
\Require Sparse Patch-based VAE decoder $\mathcal{D}_{\mathrm{sparse}}$
\Ensure High-resolution image $\mathbf{x}_{\mathrm{HR}}$
\Statex \textit{// Cascade upsampling}
\State $r \leftarrow n$
\For{$i = 1$ to $s$}
    \Statex \quad\textit{// Latent-space upsampling (replaces VAE round-trip)}
    \State $\bar{\mathbf{z}}_0^{4r} \leftarrow \mathcal{H}_\theta(\mathbf{z}_0^r)$ \Comment{Latent Hyperloop}
    \Statex
    \Statex \quad\textit{// Patchify with overlap}
    \State $\{P_j\}_{j=1}^{N_p} \leftarrow \mathrm{Patchify}(\bar{\mathbf{z}}_0^{4r})$
    \Statex
    \Statex \quad\textit{// Per-patch refinement}
    \For{$j = 1$ to $N_p$}
        \State $\bar{\mathbf{z}}_K^{(j)} \leftarrow \mathrm{DDIM\text{-}Inv}(P_j, K, \epsilon_\gamma^{\mathrm{refine}})$ \Comment{Partial inversion}
        \State $\boldsymbol{\epsilon} \sim \mathcal{N}(\mathbf{0}, \mathbf{I})$ 
        \State $\boldsymbol{\epsilon}^{\mathrm{pred}} \leftarrow \mathrm{\text{Predict Noise}}(\mathbf{z}_K^{(j)}, K, \epsilon_\gamma^{\mathrm{refine}})$ \Comment{One-step refinement}
        \State $\boldsymbol{\epsilon}^{\mathrm{inj}}\leftarrow \mathrm{Slerp}(\boldsymbol{\epsilon}^{\mathrm{pred}},\, \boldsymbol{\epsilon},\, \lambda)$ \Comment{Noise injection}
        \State $\hat{\mathbf{z}}_0^{(j)} \leftarrow \mathrm{Denoise}(\mathbf{z}_K^{(j)}, K, \boldsymbol{\epsilon}^{\mathrm{inj}})$ 
    \EndFor
    \Statex
    \Statex \quad\textit{// Patch recombination}
    \State $\mathbf{z}_0^{4r} \leftarrow \mathrm{GaussianBlend}(\{\hat{\mathbf{z}}_0^{(j)}\}_{j=1}^{N_p})$ \Comment{Smooth blending}
    \State $r \leftarrow 4r$
\EndFor
\Statex
\Statex \textit{// Stage 3: Final decode with Sparse VAE}
\State $\mathbf{x}_{\mathrm{HR}} \leftarrow \mathcal{D}_{\mathrm{sparse}}(\mathbf{z}_0^R)$ \Comment{Sparse VAE decoding}
\State \Return $\mathbf{x}_{\mathrm{HR}}$
\end{algorithmic}
\end{algorithm}
\subsection{Latent Hyperloop}
\label{app:lh_details}
We adopt a SwinIR-based architecture for the Latent Hyperloop (LH) and incorporate structural priors from the VAE decoder to guide the learned upsampling. With SDXL-Turbo refiner, LH accepts latent patch inputs with resolution $80\times80$, i.e., a $64\times64$ source crop expanded by $p=8$ context pixels on each side (input side length $s+2p$ with $s=64$), to generate upsampled latent patches with resolution $128 \times 128$. With SD-Turbo, while we could use smaller input and output patch sizes to align with SD-Turbo's convention, we find it unnecessary and use the same config as with SDXL-Turbo ($80\times80$ for input and $128 \times 128$ for output) for simple implementation. The model is trained from scratch using the AdamW optimizer with a fixed learning rate of $10^{-4}$ (no scheduling) for 20K steps on 8 NVIDIA H100 GPUs (80GB each), with a batch size of 32 per GPU (global batch size 256).
\subsection{Sparse VAE Decoding}
\label{app:sparse_vae}

\noindent\textbf{Overview.}
When decoding a high-resolution latent (spatial dimensions $\geq 512$), a single-pass VAE decode causes out-of-memory errors. The standard solution is patch-wise decoding with overlapping regions. Large overlaps produce seamless results but are slow. Our Sparse VAE Decoding enables small overlaps by ensuring consistent GroupNorm statistics across all patches.

\noindent\textbf{CrossPatchGroupNorm.}
We replace every \texttt{GroupNorm} layer in the VAE decoder with a custom \texttt{CrossPatchGroupNorm} module that can operate in two modes: (1)~standard mode, computing statistics from the current input, and (2)~cached mode, using pre-computed reference statistics.

\noindent\textbf{Reference statistics from the first batch.}
Rather than requiring a separate reference pass, we compute and cache GroupNorm statistics from the \emph{first batch} of patches. During the first batch's forward pass, a forward hook on each \texttt{CrossPatchGroupNorm} layer records the mean and variance aggregated over both the batch and spatial dimensions:
\begin{equation}
    \mu_l^{\mathrm{ref}} = \frac{1}{B \cdot |\mathcal{S}|} \sum_{b,s} \mathbf{h}_l^{(b,s)}, \quad
    \sigma_l^{2,\mathrm{ref}} = \frac{1}{B \cdot |\mathcal{S}|} \sum_{b,s} (\mathbf{h}_l^{(b,s)} - \mu_l^{\mathrm{ref}})^2,
    \label{eq:ref_stats}
\end{equation}
where $B$ is the batch size and $\mathcal{S}$ denotes the spatial and channel-group dimensions. After the first batch completes, the hooks are removed and all subsequent batches use the cached statistics automatically.

\noindent\textbf{Batched decoding.}
Patches of standard size are grouped into mini-batches and decoded in parallel. Edge patches with non-standard sizes (at image boundaries) are decoded individually with cached statistics. After all patches are decoded, adjacent patches are blended along both vertical and horizontal directions to produce the final seamless output.

\begin{lstlisting}[language=Python, caption={Sparse VAE Decoding (simplified)}]
class CrossPatchGroupNorm(nn.Module):
    """GroupNorm with optional cached stats."""
    def __init__(self, num_groups, num_channels,
                 eps=1e-5, affine=True):
        super().__init__()
        self.num_groups = num_groups
        self.num_channels = num_channels
        self.eps = eps
        self.weight = nn.Parameter(torch.ones(num_channels))
        self.bias = nn.Parameter(torch.zeros(num_channels))

        # Cached stats from reference patch
        self._ref_mean = None
        self._ref_var = None
        self._use_cached = False

    def compute_and_cache_stats(self, x):
        """
        Call this on your reference patch 
        BEFORE processing other patches.
        """
        B, C, H, W = x.shape
        G = self.num_groups
        xg = x.view(B, G, C // G, H, W)
        # Aggregate over batch + spatial dims
        self._ref_mean = xg.mean(
            dim=[0, 2, 3, 4], keepdim=True)
        self._ref_var = xg.var(
            dim=[0, 2, 3, 4], keepdim=True,
            unbiased=False)
        self._use_cached = True

    def forward(self, x):
        B, C, H, W = x.shape
        G = self.num_groups
        xg = x.view(B, G, C // G, H, W)
        if self._use_cached and self._ref_mean is not None:
            mean, var = self._ref_mean, self._ref_var
        else:
            # Normal GroupNorm behavior (for reference patch itself)
            mean = xg.mean(dim=[2,3,4], keepdim=True)
            var = xg.var(dim=[2,3,4], keepdim=True,
                         unbiased=False)
        xn = (xg - mean) / (var + self.eps).sqrt()
        xn = xn.view(B, C, H, W)
        return xn * self.weight.view(1,C,1,1) \
             + self.bias.view(1,C,1,1)

def replace_groupnorm(model):
    """Replace GN layers"""
    for name, mod in model.named_children():
        if isinstance(mod, nn.GroupNorm):
            new = CrossPatchGroupNorm(
                mod.num_groups, mod.num_channels,
                mod.eps, mod.affine)
            new.weight.data = mod.weight.data.clone()
            new.bias.data = mod.bias.data.clone()
            setattr(model, name, new)
        else:
            replace_groupnorm(mod)
\end{lstlisting}

\noindent\textbf{Decoding procedure.}
The full Sparse VAE Decoding proceeds as follows:
\begin{enumerate}
    \item Replace all GroupNorm layers in the VAE decoder with \texttt{CrossPatchGroupNorm}. 
    \item Tile the latent into overlapping patches; group standard-size patches into batches.
    \item Decode the \emph{first batch} with forward hooks that call \texttt{compute\_and\_cache\_stats} on each layer's input, then remove hooks.
    \item Decode remaining batches and edge patches; cached statistics are used automatically.
    \item Blend all decoded patches along vertical and horizontal overlap regions.
\end{enumerate}
This approach avoids a separate reference pass entirely: the first batch of patches serves as the reference, and all subsequent patches benefit from consistent normalization with zero additional overhead.

\section{Additional Quantitative Results}
\label{app:quantitative}

\subsection{Hybrid Noise Inversion}
\label{app:hybrid_noise}
InvSR\footnote{Z. Yue, K. Liao, and C. C. Loy, ``Arbitrary-steps image super-resolution via diffusion inversion,'' in \emph{CVPR}, pp. 23153--23163, 2025.} is a pretrained model for diffusion inversion, targeting the super-resolution task. It takes a coarsely upsampled latent as input and predicts its noise residual, which is added to the input latent to form the noisy latent to produce the super-resolution outcome via one diffusion forward step. This formulation is naturally compatible with our refinement stage: we can introduce a parallel InvSR branch alongside the existing DDIM inversion branch, blending their outputs to combine the structural fidelity of DDIM inversion with the detail-aware noise prediction of InvSR.

Specifically, we compute both the DDIM-inverted noise $\bar{\mathbf{z}}_K^{\mathrm{DDIM}}$ and the InvSR-predicted noise $\bar{\mathbf{z}}_K^{\mathrm{InvSR}}$, then combine them via spherical linear interpolation (Slerp):
\begin{equation}
    \mathbf{z}_K = \mathrm{Slerp}(\bar{\mathbf{z}}_K^{\mathrm{DDIM}},\; \bar{\mathbf{z}}_K^{\mathrm{InvSR}},\; \alpha), \quad \alpha = 0.8,
    \label{eq:hybrid_slerp}
\end{equation}
where $\alpha = 0.8$ assigns the dominant weight to DDIM inversion. We use Slerp rather than linear interpolation because noisy latents in diffusion models lie on a high-dimensional hypersphere, and Slerp preserves the norm and angular distribution of the interpolated noise.

Table~\ref{tab:invsr} shows that InvSR alone does not provide strong scores when generating high-resolution images. However, it is highly compatible with our pipeline, and incorporating InvSR into \methodname{} consistently improves both FID and IS across resolutions. At 2K, FID improves from 50.02 to 49.33 with a corresponding IS increase from 14.03 to 14.24. At 4K, FID drops from 52.38 to 51.34. These gains come at the cost of one additional InvSR forward pass per patch, which adds moderate overhead but can be worthwhile when quality is prioritized over speed.

\begin{table}[h]
    \centering
    \small
    \caption{\textbf{Hybrid noise inversion with InvSR.} Slerp blending ($\alpha\!=\!0.8$ DDIM, $0.2$ InvSR) consistently improves quality.}
    \label{tab:invsr}
    \begin{tabular}{lcccccc}
        \toprule
        & \multicolumn{3}{c}{$2048 \times 2048$} & \multicolumn{3}{c}{$4096 \times 4096$} \\
        \cmidrule(lr){2-4} \cmidrule(lr){5-7}
        \textbf{Config} & FID $\downarrow$ & IS $\uparrow$ &Time & FID $\downarrow$ & IS $\uparrow$ &Time\\
        \midrule
        \rowcolor{gray!15} InvSR (only)            & 65.32 & 13.22 & - & 76.87 & 12.78 & -\\
        \midrule
        \methodname{}            & 50.02 & 14.03 &\textbf{2.78} & 52.38 & 13.99 &\textbf{8.04}\\
        + InvSR    & \textbf{49.33} & \textbf{14.24} &5.12 & \textbf{51.34} & \textbf{14.04} &13.76 \\
        \bottomrule
    \end{tabular}
\end{table}
\subsection{Ablation Study}
\begin{figure*}[t]
\centering

\newcommand{\subfigtitle}[1]{%
    \vspace{1mm}
    {\small\bfseries #1}
    \vspace{1mm}
}

\begin{tabular}{@{}p{0.62\textwidth}@{\hspace{0.035\textwidth}}p{0.34\textwidth}@{}}

\begin{minipage}[t]{\linewidth}
\vspace{0pt}
\centering
\scriptsize
\renewcommand{\arraystretch}{1.12}
\setlength{\tabcolsep}{4pt}

\resizebox{\linewidth}{!}{%
\begin{tabular}{lccc}
\toprule
\textbf{Variant} 
& \textbf{FID} $\downarrow$ 
& \textbf{IS} $\uparrow$ 
& \textbf{Latency (s)} $\downarrow$ \\
\midrule
\basename{} (SD-Turbo) 
& 56.50 & 13.42 & 16.37 \\
\quad + Latent Hyperloop$^{*}$ 
& 59.18 & 13.12 & 11.53 \\
\quad + Latent Hyperloop 
& \textbf{51.99} & 13.89 & 11.83 \\
\rowcolor{green!8}
\quad + Sparse Decoding 
& 52.38 & \textbf{13.99} & \textbf{8.04} \\
\bottomrule
\end{tabular}
}
{\scriptsize $^{*}$ excludes the context-aware patch-based mechanism.}

\vspace{1mm}
\subfigtitle{(a) Component-wise Ablation on 4K Generation}

\vspace{4mm}

\resizebox{\linewidth}{!}{%
\begin{tabular}{llcc}
\toprule
\textbf{Base Generator} 
& \textbf{Refinement Model} 
& \textbf{FID} $\downarrow$ 
& \textbf{IS} $\uparrow$ \\
\midrule
SDXL~\citep{podell2023sdxl} 
& SD-Turbo~\citep{sauer2024adversarial} 
& 52.38 & 13.99 \\
FLUX~\citep{flux2024} 
& SD-Turbo~\citep{sauer2024adversarial} 
& 51.75 & \textbf{16.61} \\
SDXL~\citep{podell2023sdxl} 
& SDXL-Turbo~\citep{sauer2024adversarial} 
& \underline{51.32} & 13.93 \\
FLUX~\citep{flux2024} 
& SDXL-Turbo~\citep{sauer2024adversarial} 
& \textbf{50.87} & \underline{16.52} \\
SDXL~\citep{podell2023sdxl} 
& PixArt-$\Sigma$~\citep{chen2024pixartsigma} 
& 53.47 & 13.70 \\
\bottomrule
\end{tabular}
}\label{tab:ablation_backbone}
\subfigtitle{(b) Robustness Across Backbone Combinations}

\end{minipage}
&
\begin{minipage}[t]{\linewidth}
\vspace{0pt}
\centering

\includegraphics[width=\linewidth]{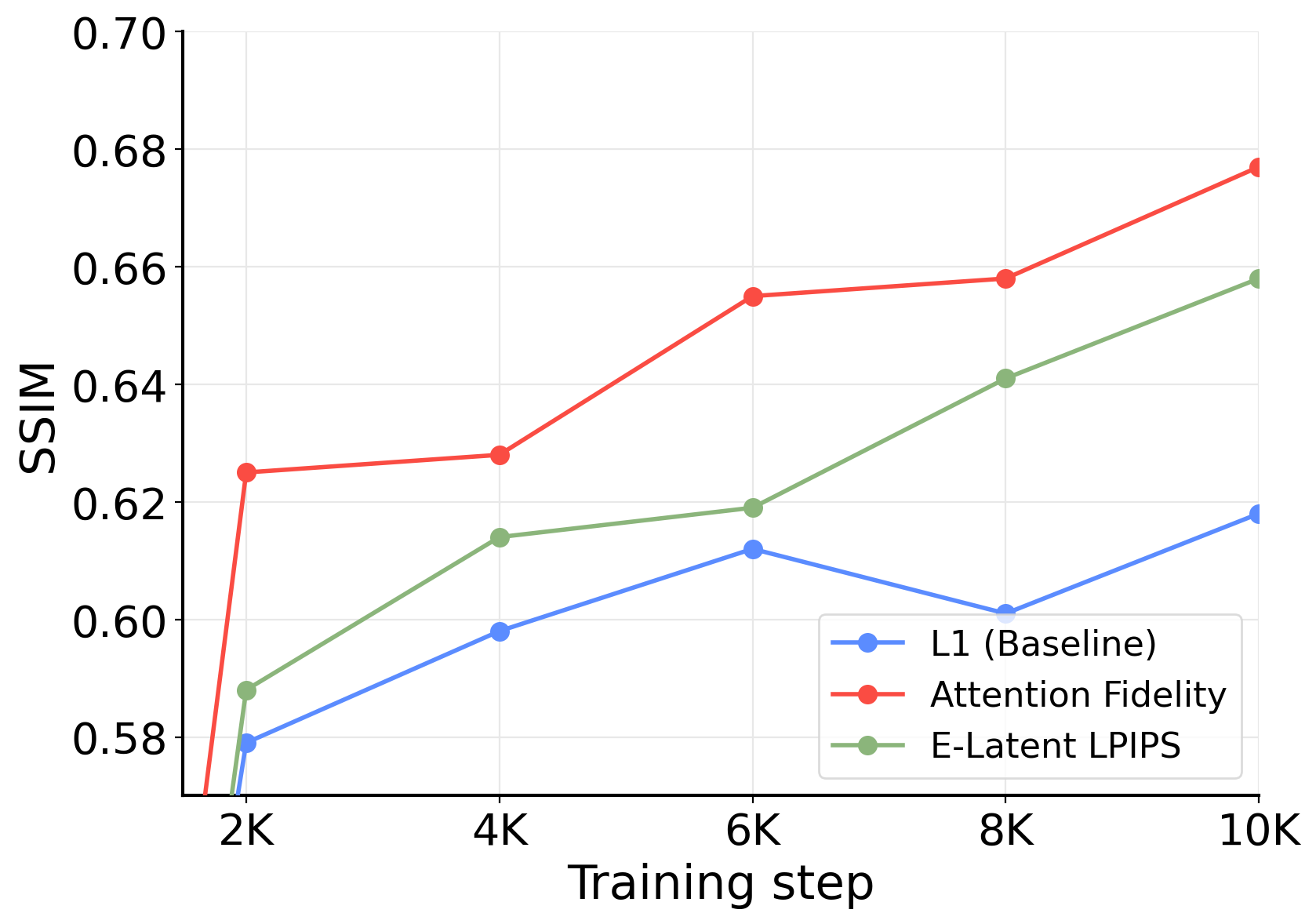}
\subfigtitle{(c) Objective Convergence}

\vspace{2mm}

\includegraphics[width=\linewidth]{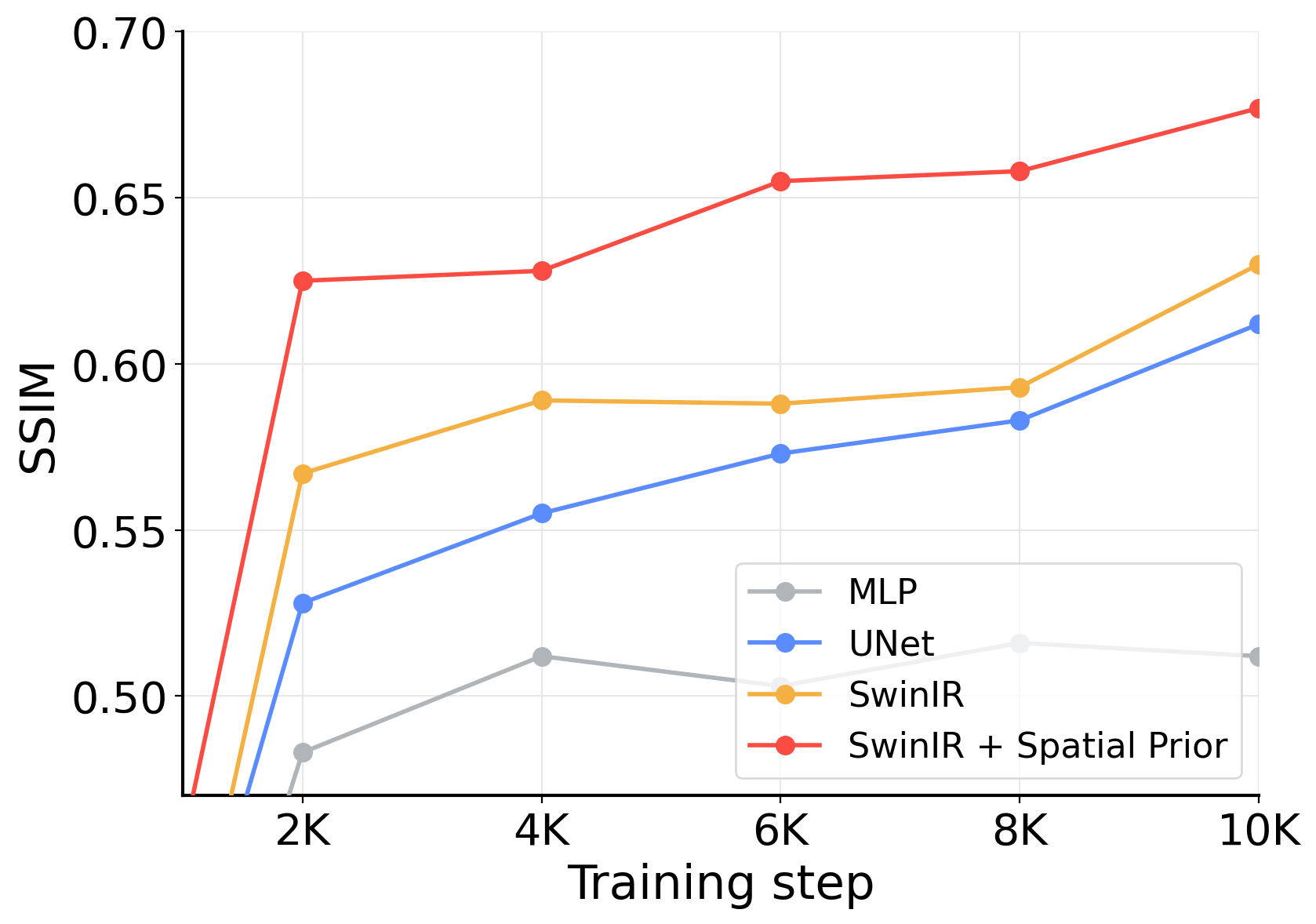}
\subfigtitle{(d) Architecture Convergence}

\end{minipage}

\end{tabular}
\caption{
\textbf{Latent Hyperloop ablations and convergence analysis.}
\textbf{Left:} We report 4K generation ablations and robustness across different base/refinement backbone combinations. Latent Hyperloop improves generation quality and latency, while Sparse Patch-based Decoding further reduces inference time with only a small FID trade-off. 
\textbf{Right:} SSIM convergence curves. (c) Attention-based objectives accelerate convergence and improve the final reconstruction quality over the $\ell_1$ baseline and E-LatentLPIPS~\citep{kang2024distilling}; (d) the Spatial Prior architecture consistently outperforms alternative designs.
}
\label{fig:latent_hyperloop_ablation_main}
\end{figure*}

\minisection{Why Each Component Matters.} Fig.~\ref{fig:latent_hyperloop_ablation_main}a reveals why both components are necessary. First, we observe that a naive implementation of the Latent Hyperloop (marked with $^*$) successfully reduces latency but causes a clear drop in fidelity (FID degrades from 56.50 to 59.18) due to lack of boundary context. However, by incorporating our \textbf{\textit{context-expanded mechanism}}, the full Latent Hyperloop not only resolves these artifacts but improves FID to 51.99 while reducing latency by about 28\% relative to the baseline (16.37s $\to$ 11.83s). Subsequently, \textbf{Sparse Patch-based Decoding} provides additive efficiency gains by reducing redundancy in the decoding and sampling stages. Together, the two components reduce the total inference time to 8.04s -- a \textbf{2$\times$} acceleration compared to \basename{} (16.37s) -- while improving visual quality.

\minisection{Robustness to Model Configurations.}
To assess the versatility of \methodname{}, we evaluate its performance across diverse settings, varying both the base model and the refinement module. As shown in Fig.~\ref{fig:latent_hyperloop_ablation_main}b, \methodname{} consistently produces strong FID and IS scores regardless of the underlying model.

\minisection{Latent Hyperloop Design.}
Fig.~\ref{fig:latent_hyperloop_ablation_main}c--d analyze the design choices behind \textbf{Latent Hyperloop}. In Fig.~\ref{fig:latent_hyperloop_ablation_main}c, Attention Alignment improves convergence and final decoded-image fidelity over both the $\ell_1$ baseline and E-LatentLPIPS~\citep{kang2024distilling}, a perceptual loss computed in latent space, showing that refiner attention provides useful structure-aware supervision beyond latent reconstruction alone. In Fig.~\ref{fig:latent_hyperloop_ablation_main}d, \textbf{Spatial Prior} further improves SSIM by reusing spatial expansion behavior already present in the VAE decoder. Together, these ablations justify the final \textbf{Latent Hyperloop} design.

\minisection{Context Size.}
\label{app:context_ablation}
Table~\ref{tab:context_ablation} varies the number of context pixels $p$ added on each side of the $64\times64$ source crop, i.e., the input side length $s+2p$ of Latent Hyperloop (Sec.~\ref{sec:latent_hyperloop}). Quality improves sharply as context is added, with FID decreasing from 59.18 at $p=0$ to 51.99 at $p=8$, after which the returns diminish: increasing $p$ to 12 or 16 changes FID by less than 0.35. We therefore use $p=8$ as the trade-off between quality and cost, since a larger context adds computation for marginal gain.

\begin{table}[h]
    \centering
    \small
    \caption{\textbf{Ablation on the context size $p$ of Latent Hyperloop.} Same 4K setting as Fig.~\ref{fig:latent_hyperloop_ablation_main}a (SD-Turbo refiner, before adding Sparse Patch-based Decoding); $p=0$ corresponds to the variant without context expansion.}
    \label{tab:context_ablation}
    \begin{tabular}{lccccc}
        \toprule
        Context pixels $p$ & 0 & 4 & 8 (ours) & 12 & 16 \\
        \midrule
        FID $\downarrow$ & 59.18 & 55.86 & 51.99 & 51.67 & 51.71 \\
        IS $\uparrow$ & 13.12 & 13.48 & 13.89 & 13.82 & 13.89 \\
        \bottomrule
    \end{tabular}
\end{table}

\begin{figure}[t]
    \centering
    \includegraphics[width=0.6\linewidth]{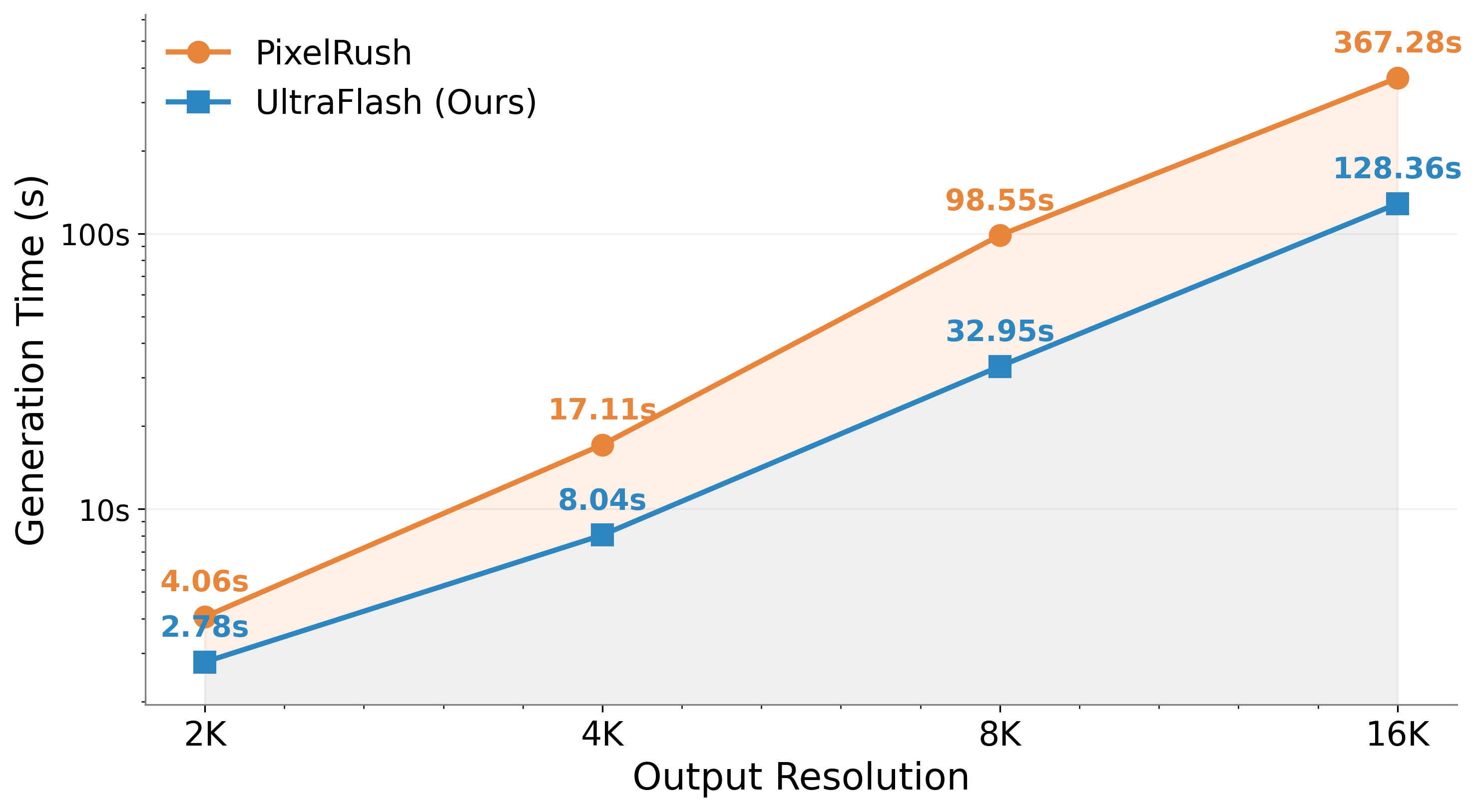}
    \caption{
    \textbf{Scalability Analysis.} 
    Comparison of inference latency across increasing resolutions (log scale). \methodname{} (blue) offers significantly better scaling properties than \basename{} (orange). 
    }
    \label{fig:speed_comparison}
\end{figure}
\section{Additional Qualitative Results}
\label{app:qualitative}

\subsection{Latent Hyperloop vs. Image Bilinear Interpolation}
\label{app:lba_visual}

\begin{figure*}[t]
    \centering
    \includegraphics[width=1.0\textwidth]{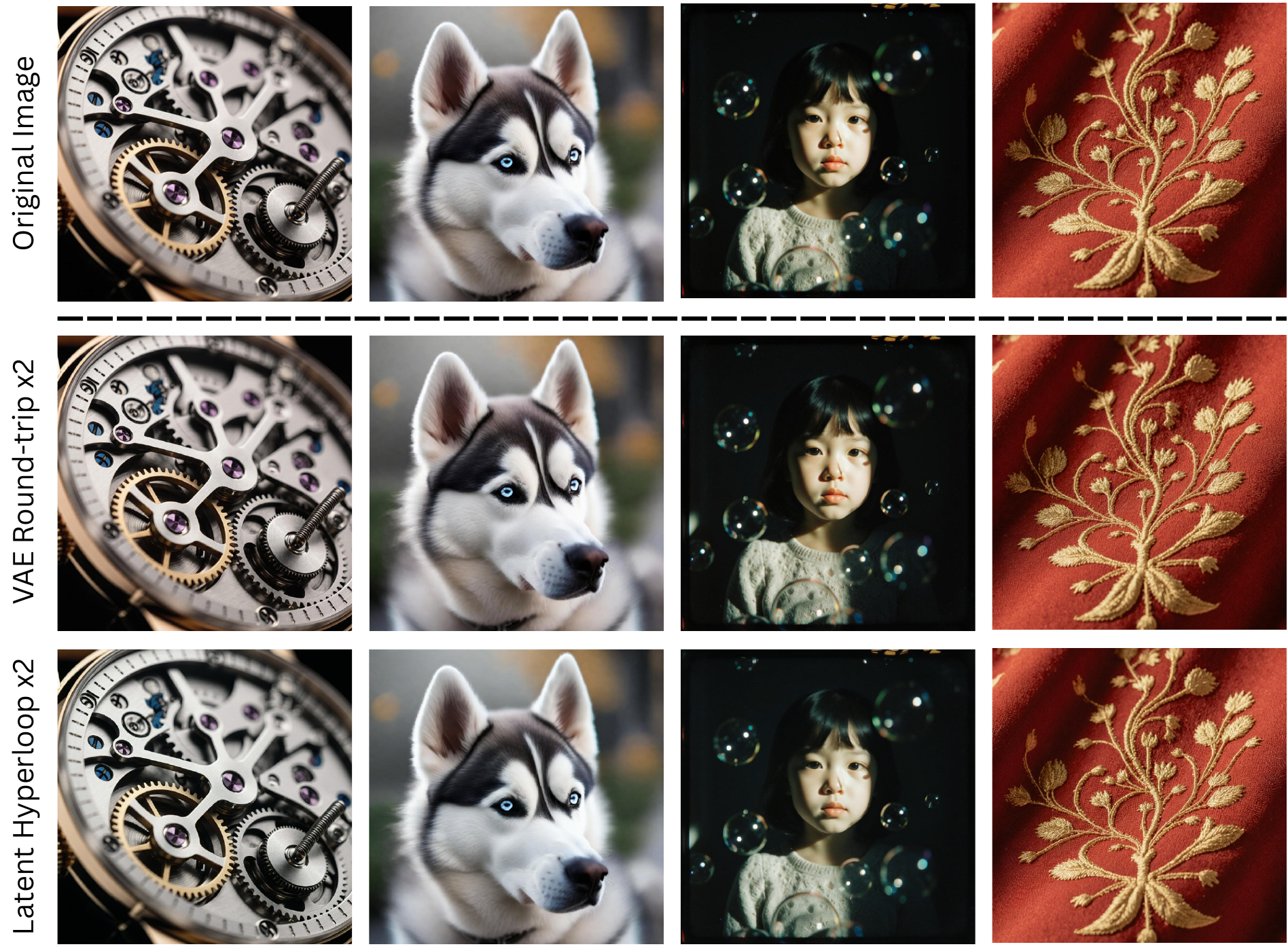}
    \caption{\textbf{Latent Hyperloop vs.\ image bilinear interpolation.} Our Latent Hyperloop produces outputs that are visually indistinguishable from the expensive pixel-space round-trip. Best viewed \textbf{zoomed-in}.}
    \label{fig:lba_comparison}
\end{figure*}

Fig.~\ref{fig:lba_comparison} compares the output of our Latent Hyperloop adapter against the standard pixel-space pipeline (VAE decode $\to$ bilinear interpolation $\to$ VAE encode). For each input latent, we decode both versions to pixel space. The Latent Hyperloop closely matches the pixel-space ground truth across diverse content types.

\subsection{Effect of Training Objectives on Latent Hyperloop Reconstruction}
\label{app:loss_visual}

\begin{figure*}[t]
    \centering
    \includegraphics[width=0.98\textwidth]{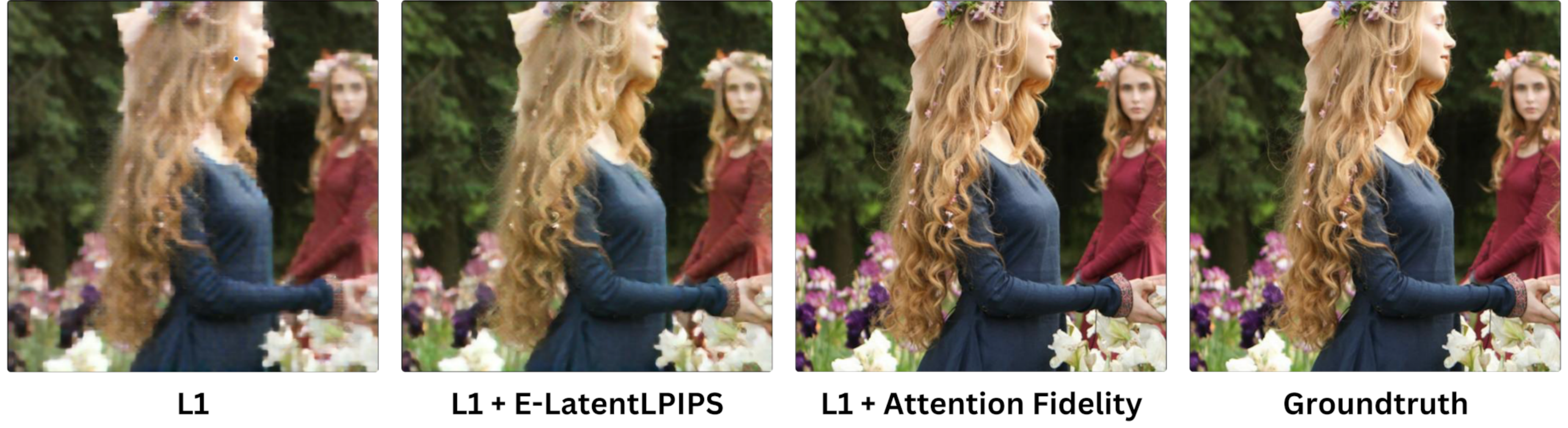}
    \caption{\textbf{Effect of training objectives on Latent Hyperloop reconstruction.} From left to right: reconstructions trained with $\ell_1$ only, $\ell_1$ + E-LatentLPIPS, and $\ell_1$ + Attention Alignment, followed by the ground-truth reconstruction. All variants use the same latent upsampling backbone and are trained under the same setting. While $\ell_1$ preserves the overall structure, it produces noticeably smooth reconstructions, and E-LatentLPIPS provides only marginal visual improvement. The proposed Attention Alignment objective better preserves local structure and fine facial details, yielding reconstructions that are visually closest to the ground truth. Best viewed zoomed in.}
    \label{fig:loss}
\end{figure*}

Fig.~\ref{fig:loss} provides a qualitative comparison of the training objectives used for learning the Latent Hyperloop adapter. Using only the $\ell_1$ reconstruction loss preserves the coarse object layout, but the reconstructed image remains visibly over-smoothed, with weakened local contrast and softened facial structure. Adding E-LatentLPIPS yields only limited perceptual improvement over $\ell_1$ alone. In contrast, incorporating the proposed Attention Alignment objective produces substantially sharper and more faithful reconstructions, especially in fine-grained regions such as the eyes, nose contour, and surrounding fur. These visual differences support the design choice of using Attention Alignment to improve latent-space upsampling quality.

\subsection{Additional \methodname{} Outputs}
\label{app:additional_outputs}

\begin{figure*}[t]
    \centering
    \includegraphics[width=0.98\textwidth]{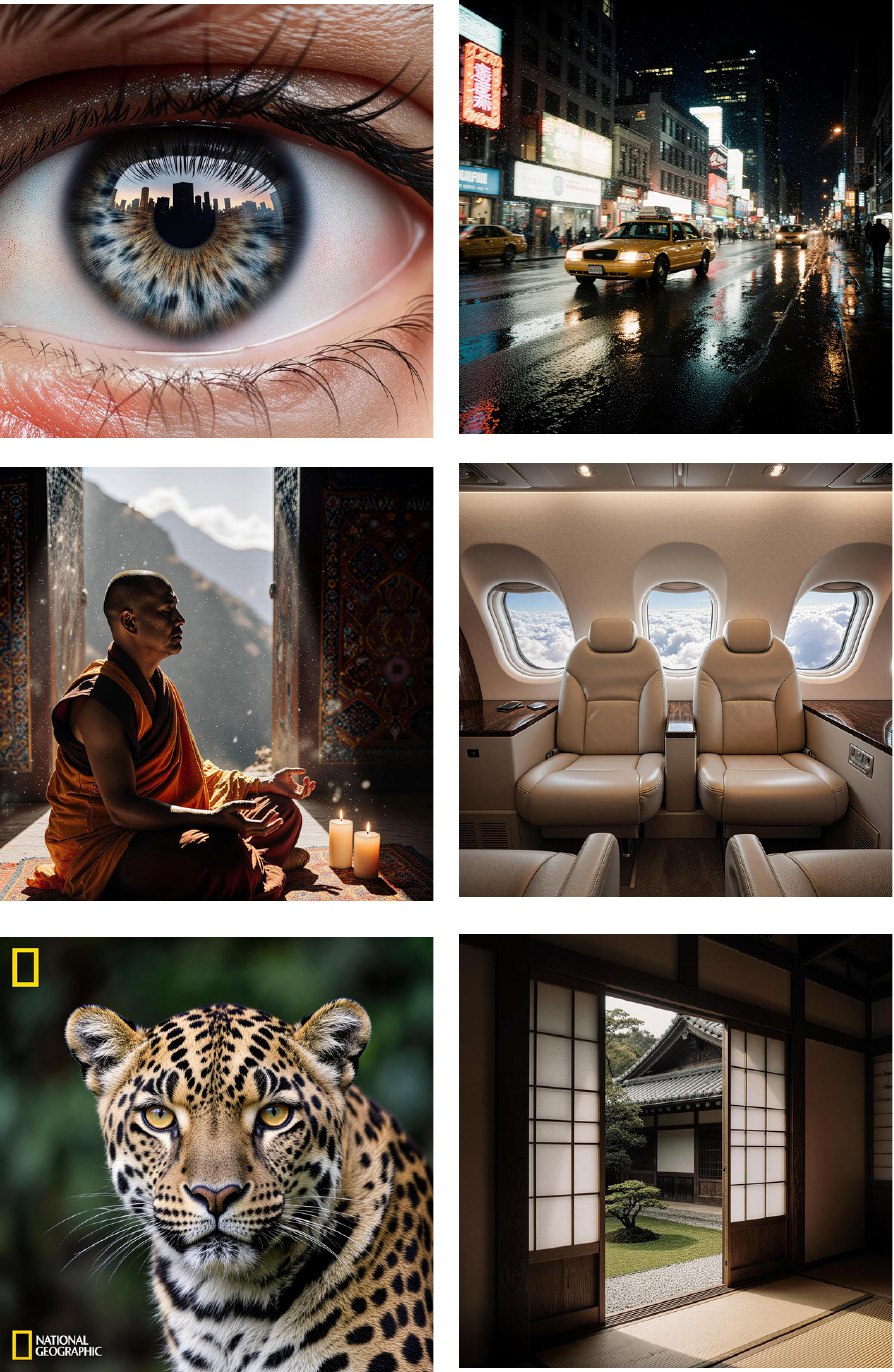}
    \caption{\textbf{Uncurated \methodname{} outputs at $4096\times4096$.} Generated with FLUX base model and \methodname{} refinement pipeline.}
    \label{fig:supp_flux_qual1}
\end{figure*}

\begin{figure*}[t]
    \centering
    \includegraphics[width=0.98\textwidth]{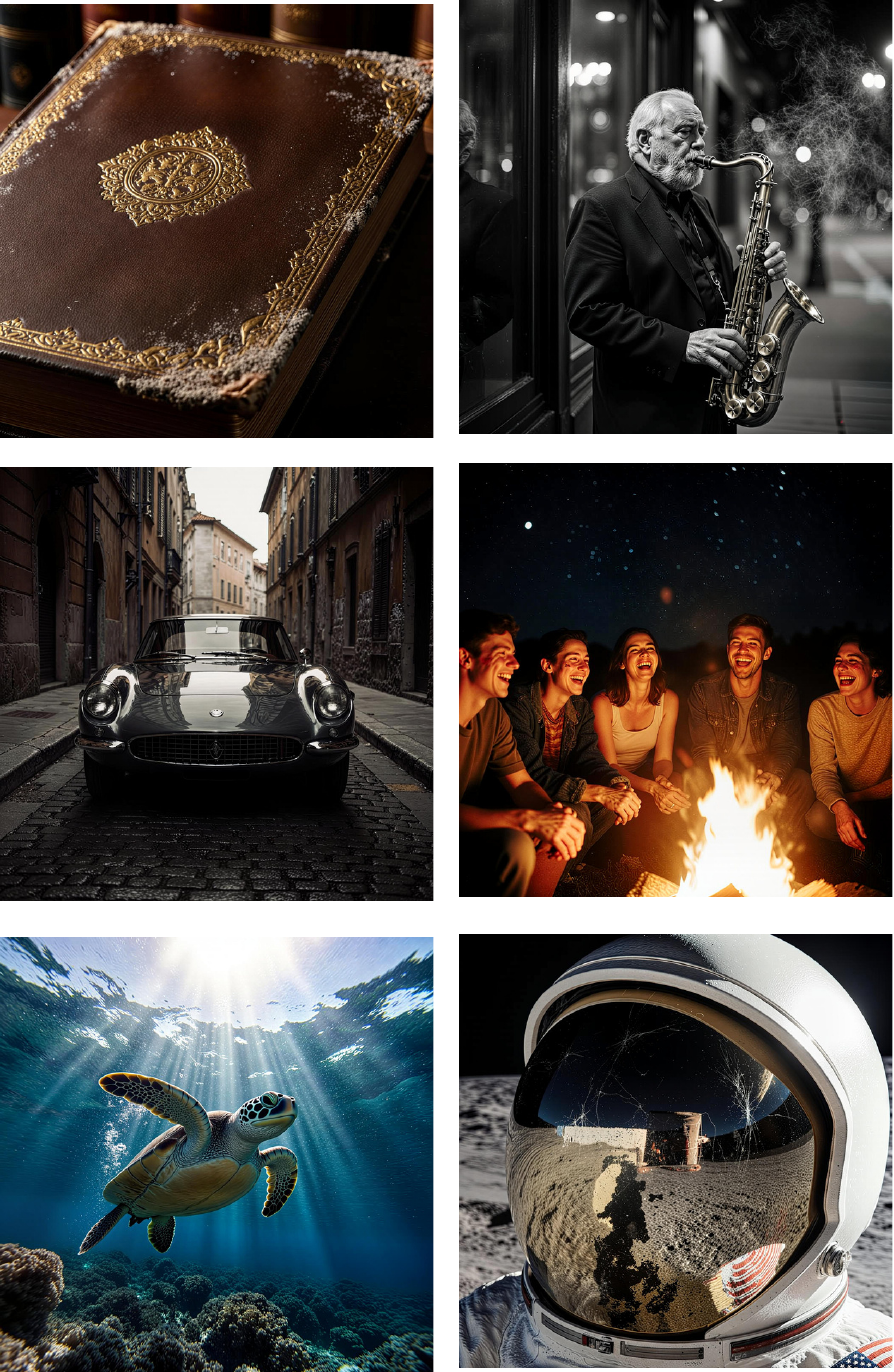}
    \caption{\textbf{Uncurated \methodname{} outputs at $4096\times4096$.} Generated with FLUX base model and \methodname{} refinement pipeline.}
    \label{fig:supp_flux_qual2}
\end{figure*}

\begin{figure*}[t]
    \centering
    \includegraphics[width=0.98\textwidth]{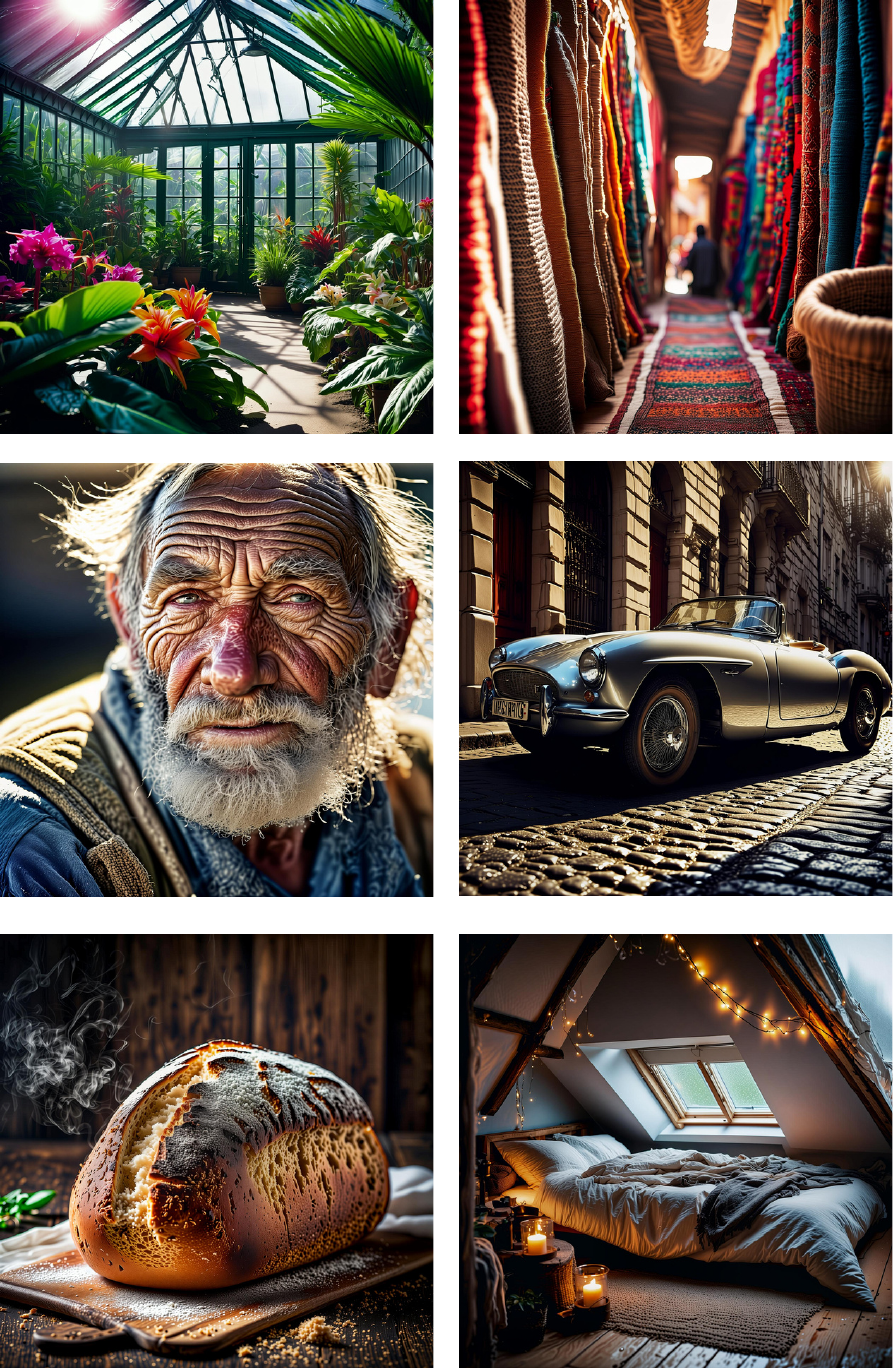}
    \caption{\textbf{Uncurated \methodname{} outputs at $4096\times4096$.} Generated with SDXL base model and \methodname{} refinement pipeline.}
    \label{fig:supp_sdxl_qual1}
\end{figure*}

\begin{figure*}[t]
    \centering
    \includegraphics[width=0.98\textwidth]{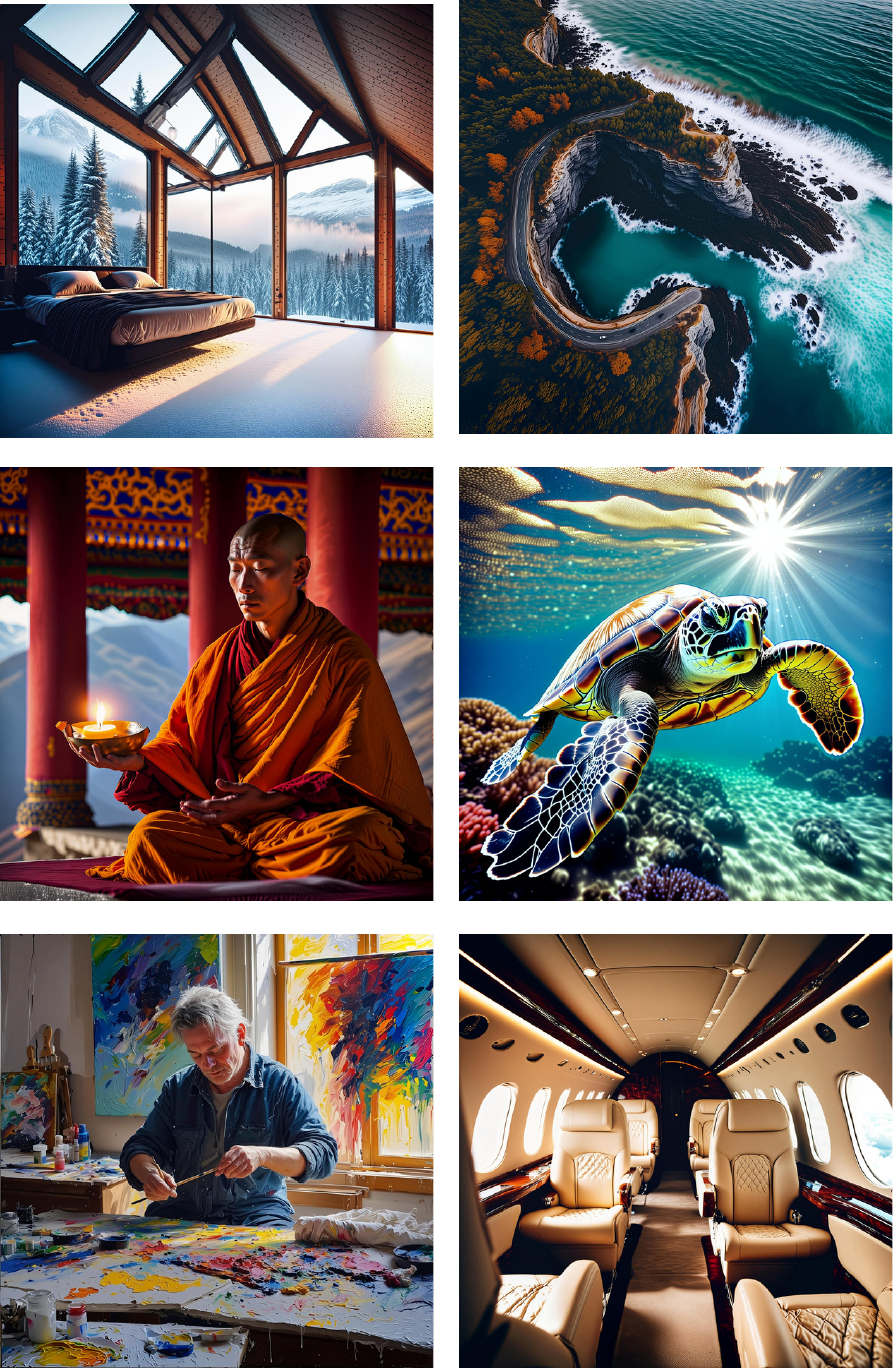}
    \caption{\textbf{Uncurated \methodname{} outputs at $4096\times4096$.} Generated with SDXL base model and \methodname{} refinement pipeline.}
    \label{fig:supp_sdxl_qual2}
\end{figure*}

Figs.~\ref{fig:supp_flux_qual1} and~\ref{fig:supp_flux_qual2} present additional $4096\times4096$ outputs using FLUX as the base model, while Figs.~\ref{fig:supp_sdxl_qual1} and~\ref{fig:supp_sdxl_qual2} show results using SDXL as the base model. All images are generated with \methodname{} refinement on a single A100 GPU, demonstrating that our pipeline generalizes across different base architectures.

\section{Limitations}
\label{app:limitations}

UltraFlash is developed and evaluated for cascade-based megapixel image synthesis. This focus allows us to isolate and optimize a concrete bottleneck that appears in practical high-resolution text-to-image pipelines: VAE-heavy transitions between latent and pixel space. We therefore view the method as a targeted acceleration strategy for image synthesis rather than a claim about all generative settings. Other modalities, such as video generation, introduce additional temporal and architectural considerations that are outside the scope of this work. At the same time, the central observation behind UltraFlash is broader: as denoising becomes faster, the surrounding transition costs become increasingly important. Exploring transition-aware acceleration beyond image cascades is a natural direction for future work.

\section{Broader Impact}
\label{app:broader-impact}

UltraFlash improves the efficiency of high-resolution image synthesis by reducing transition overhead in cascade pipelines. This can make megapixel generation more practical for creative tools, design workflows, data visualization, and research settings where high-resolution outputs are useful but computational cost is a barrier. At the same time, faster and higher-resolution generation should be deployed with the same safeguards used for the underlying generative models, including content filtering, provenance or watermarking mechanisms when available, and responsible-use policies. UltraFlash does not introduce a new generative prior or bypass the safety behavior of the base models; it accelerates the transition operations around existing cascade refinement. We therefore view its impact as primarily improving efficiency, while inheriting the broader benefits and risks of the image-generation systems on which it is built.
\clearpage
\newpage
\section*{NeurIPS Paper Checklist}

\begin{enumerate}

\item {\bf Claims}
    \item[] Question: Do the main claims made in the abstract and introduction accurately reflect the paper's contributions and scope?
    \item[] Answer: \answerYes{}.
    \item[] Justification: The abstract and introduction state the paper's scope as transition-efficient megapixel synthesis, with contributions including Latent Hyperloop, Sparse Patch-based Decoding, and empirical evaluation in cascade pipelines. The claims are supported by the experiments in Section~\ref{sec:experiments}, and comparisons with native 4K and FLUX-based methods are explicitly framed as contextual because they use different backbones or training assumptions.

\item {\bf Limitations}
    \item[] Question: Does the paper discuss the limitations of the work performed by the authors?
    \item[] Answer: \answerYes{}.
    \item[] Justification: The paper discusses limitations in Appendix~\ref{app:limitations}. The discussion frames the scope of UltraFlash as cascade-based megapixel image synthesis and notes that extending transition-aware acceleration beyond image cascades is a natural direction for future work.

\item {\bf Theory assumptions and proofs}
    \item[] Question: For each theoretical result, does the paper provide the full set of assumptions and a complete (and correct) proof?
    \item[] Answer: \answerNA{}.
    \item[] Justification: The paper does not present formal theoretical results or theorems. The mathematical expressions are used to define the cascade transition operators and training objectives.

\item {\bf Experimental result reproducibility}
    \item[] Question: Does the paper fully disclose all the information needed to reproduce the main experimental results of the paper to the extent that it affects the main claims and/or conclusions of the paper (regardless of whether the code and data are provided or not)?
    \item[] Answer: \answerYes{}.
    \item[] Justification: Section~\ref{sec:experiments} describes the base models, refinement models, datasets/prompts, metrics, resolutions, hardware, and baseline settings. The method section and supplemental material specify the Latent Hyperloop architecture, training objective, implementation details, and Sparse Patch-based Decoding procedure needed to reproduce the main algorithmic results.

\item {\bf Open access to data and code}
    \item[] Question: Does the paper provide open access to the data and code, with sufficient instructions to faithfully reproduce the main experimental results, as described in supplemental material?
    \item[] Answer: \answerNo{}.
    \item[] Justification: The current submission does not include public code or released checkpoints. We plan to release code and trained adapters upon acceptance.

\item {\bf Experimental setting/details}
    \item[] Question: Does the paper specify all the training and test details (e.g., data splits, hyperparameters, how they were chosen, type of optimizer) necessary to understand the results?
    \item[] Answer: \answerYes{}.
    \item[] Justification: The experimental section reports the evaluation prompts, resolutions, metrics, hardware, base/refinement models, and main baseline configurations. The method and supplemental material describe Latent Hyperloop training data, architecture, losses, optimizer, training steps, and decoding settings.

\item {\bf Experiment statistical significance}
    \item[] Question: Does the paper report error bars suitably and correctly defined or other appropriate information about the statistical significance of the experiments?
    \item[] Answer: \answerNo{}.
    \item[] Justification: We do not report error bars because each full 4K evaluation is computationally expensive and uses 1000 prompts. To reduce variance, all quantitative comparisons use the same prompt set, metrics, hardware, and evaluation protocol.

\item {\bf Experiments compute resources}
    \item[] Question: For each experiment, does the paper provide sufficient information on the computer resources (type of compute workers, memory, time of execution) needed to reproduce the experiments?
    \item[] Answer: \answerYes{}.
    \item[] Justification: The experimental section reports that inference experiments are run on a single NVIDIA A100-40GB GPU and provides wall-clock runtime for each method. The paper also reports model size for Latent Hyperloop and the inference time needed to reproduce the main speed claims.

\item {\bf Code of ethics}
    \item[] Question: Does the research conducted in the paper conform, in every respect, with the NeurIPS Code of Ethics \url{https://neurips.cc/public/EthicsGuidelines}?
    \item[] Answer: \answerYes{}.
    \item[] Justification: The work uses existing pretrained image-generation models and public datasets/prompts for research evaluation. We do not collect sensitive personal data or deploy the system, and we discuss potential misuse and broader impacts of improved high-resolution generation in Appendix~\ref{app:broader-impact}.

\item {\bf Broader impacts}
    \item[] Question: Does the paper discuss both potential positive societal impacts and negative societal impacts of the work performed?
    \item[] Answer: \answerYes{}.
    \item[] Justification: The paper discusses broader impacts in Appendix~\ref{app:broader-impact}, including efficiency benefits for high-resolution image synthesis and the need to retain safeguards such as content filtering, provenance or watermarking mechanisms when available, and responsible-use policies inherited from the underlying generative models.

\item {\bf Safeguards}
    \item[] Question: Does the paper describe safeguards that have been put in place for responsible release of data or models that have a high risk for misuse (e.g., pre-trained language models, image generators, or scraped datasets)?
    \item[] Answer: \answerYes{}.
    \item[] Justification: The proposed modules are lightweight transition operators rather than standalone image generators. As discussed in Appendix~\ref{app:broader-impact}, UltraFlash does not introduce a new generative prior or bypass the safety behavior of the underlying generative models; it should be used with the same content filtering, provenance or watermarking mechanisms when available, and responsible-use policies as the base models. If released, the code and adapters will be distributed for research use with documentation, and no new scraped image dataset is released.

\item {\bf Licenses for existing assets}
    \item[] Question: Are the creators or original owners of assets (e.g., code, data, models), used in the paper, properly credited and are the license and terms of use explicitly mentioned and properly respected?
    \item[] Answer: \answerYes{}.
    \item[] Justification: The paper cites the pretrained models, datasets, and baseline methods used in the experiments, including SDXL, SDXL-Turbo, SD-Turbo, JourneyDB, LAION-Aesthetics, FouriScale, DemoFusion, FreeScale, and \basename{}. We use these assets according to their intended research/evaluation settings and provide references for attribution.

\item {\bf New assets}
    \item[] Question: Are new assets introduced in the paper well documented and is the documentation provided alongside the assets?
    \item[] Answer: \answerNA{}.
    \item[] Justification: The paper does not introduce a new dataset or benchmark. The proposed Latent Hyperloop adapter and decoding procedure are documented in the method and supplemental material; if released, the trained adapters and code will include documentation and usage instructions.

\item {\bf Crowdsourcing and research with human subjects}
    \item[] Question: For crowdsourcing experiments and research with human subjects, does the paper include the full text of instructions given to participants and screenshots, if applicable, as well as details about compensation (if any)?
    \item[] Answer: \answerNo{}.
    \item[] Justification: The user preference study was an unpaid, voluntary perceptual evaluation. Participants were presented with side-by-side generated images and asked to select the one with better overall quality, text alignment, and structure. Because no commercial crowdsourcing platform was used and no financial compensation was provided, platform and compensation details are not applicable; the current submission does not include full interface screenshots.

\item {\bf Institutional review board (IRB) approvals or equivalent for research with human subjects}
    \item[] Question: Does the paper describe potential risks incurred by study participants, whether such risks were disclosed to the subjects, and whether Institutional Review Board (IRB) approvals (or an equivalent approval/review based on the requirements of your country or institution) were obtained?
    \item[] Answer: \answerNo{}.
    \item[] Justification: The user study is a low-risk anonymous preference evaluation of generated images, but the current submission does not describe IRB or equivalent review. We will add details on participant risk, consent, anonymity, and institutional requirements in the supplemental material.

\item {\bf Declaration of LLM usage}
    \item[] Question: Does the paper describe the usage of LLMs if it is an important, original, or non-standard component of the core methods in this research? Note that if the LLM is used only for writing, editing, or formatting purposes and does \emph{not} impact the core methodology, scientific rigor, or originality of the research, declaration is not required.
    \item[] Answer: \answerNA{}.
    \item[] Justification: The core method does not use LLMs as an algorithmic component. Any language-model assistance, if used, was limited to writing, editing, or formatting and did not affect the scientific method, experiments, or originality of the work.

\end{enumerate}

\end{document}